\documentclass[10pt,twocolumn,letterpaper]{article}

\usepackage{iccv}
\usepackage{times}
\usepackage{epsfig}
\usepackage{graphicx}
\usepackage{amsmath}
\usepackage{amssymb}
\usepackage{subfig}
\usepackage{algorithm}
\usepackage[noend]{algpseudocode}
\usepackage{multirow}
\usepackage[toc,page]{appendix}
\usepackage{stfloats}

\usepackage[breaklinks=true,bookmarks=false]{hyperref}

\iccvfinalcopy 

\def\iccvPaperID{****} 

\begin{document}

\title{Weakly Supervised Object Detection with Segmentation Collaboration}

\author{Xiaoyan Li$^{1,2}$\hspace{25pt}Meina Kan$^1$\hspace{25pt}Shiguang Shan$^{1,2,3}$\hspace{25pt}Xilin Chen$^{1,2}$\\
$^1$Key Lab of Intelligent Information Processing of Chinese Academy of Sciences (CAS), \\
Institute of Computing Technology, CAS, Beijing 100190, China\\
$^2$University of Chinese Academy of Sciences, Beijing 100049, China\\
$^3$CAS Center for Excellence in Brain Science and Intelligence Technology, Shanghai 200031, China\\
{\tt\small xiaoyan.li@vipl.ict.ac.cn\hspace{25pt}\{kanmeina, sgshan, xlchen\}@ict.ac.cn}
}

%
%
%
%


\maketitle

\begin{abstract}
    Weakly supervised object detection aims at learning precise object detectors, given image category labels. In recent prevailing works, this problem is generally formulated as a multiple instance learning module guided by an image classification loss. The object bounding box is assumed to be the one contributing most to the  classification among all proposals. However, the region contributing most is also likely to be a crucial part or the supporting context of an object. To obtain a more accurate detector, in this work we propose a novel end-to-end weakly supervised detection approach, where a newly introduced generative adversarial segmentation module interacts with the conventional detection module in a collaborative loop.  
    The collaboration mechanism takes full advantages of the complementary interpretations of the weakly supervised localization task, namely detection and segmentation tasks, forming a more comprehensive solution. Consequently,  our method obtains more precise object bounding boxes, rather than parts or irrelevant surroundings. Expectedly, the proposed method achieves an accuracy of 51.0\% on the PASCAL VOC 2007 dataset, outperforming the state-of-the-arts and demonstrating its superiority for weakly supervised object detection. 
\end{abstract}

\section{Introduction}

As the data-driven approaches prevail on object detection task in both academia and industry, the amount of data in an object detection benchmark is expected to be larger and larger. However, annotating object bounding boxes is both costly and time-consuming. In order to reduce the labeling workload, researchers hope to make object detectors work in a weakly-supervised fashion, \eg learning a detector with only category labels rather than bounding boxes.
   
 
 \begin{figure}[t]
	\setlength{\abovecaptionskip}{-0.2cm}
	\setlength{\belowcaptionskip}{-0.cm}
	\vspace{-18pt}	
	\begin{center}
		\subfloat[][Previous works]{
        			\includegraphics[height=1.6in]{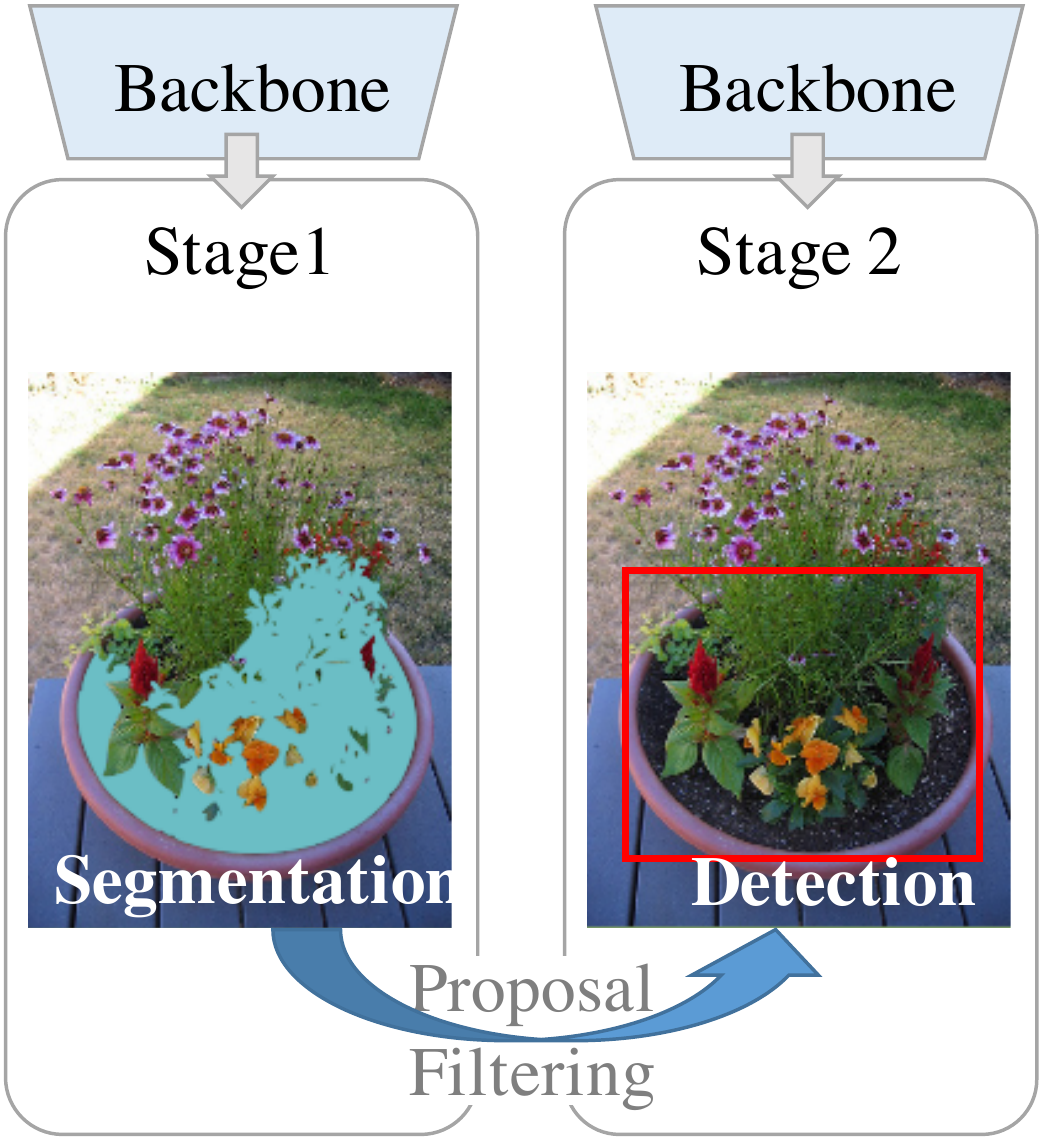}
        			\label{fig:pre-pipeline}} \hspace{10pt}
		\subfloat[][Ours]{
        			\includegraphics[height=1.6in]{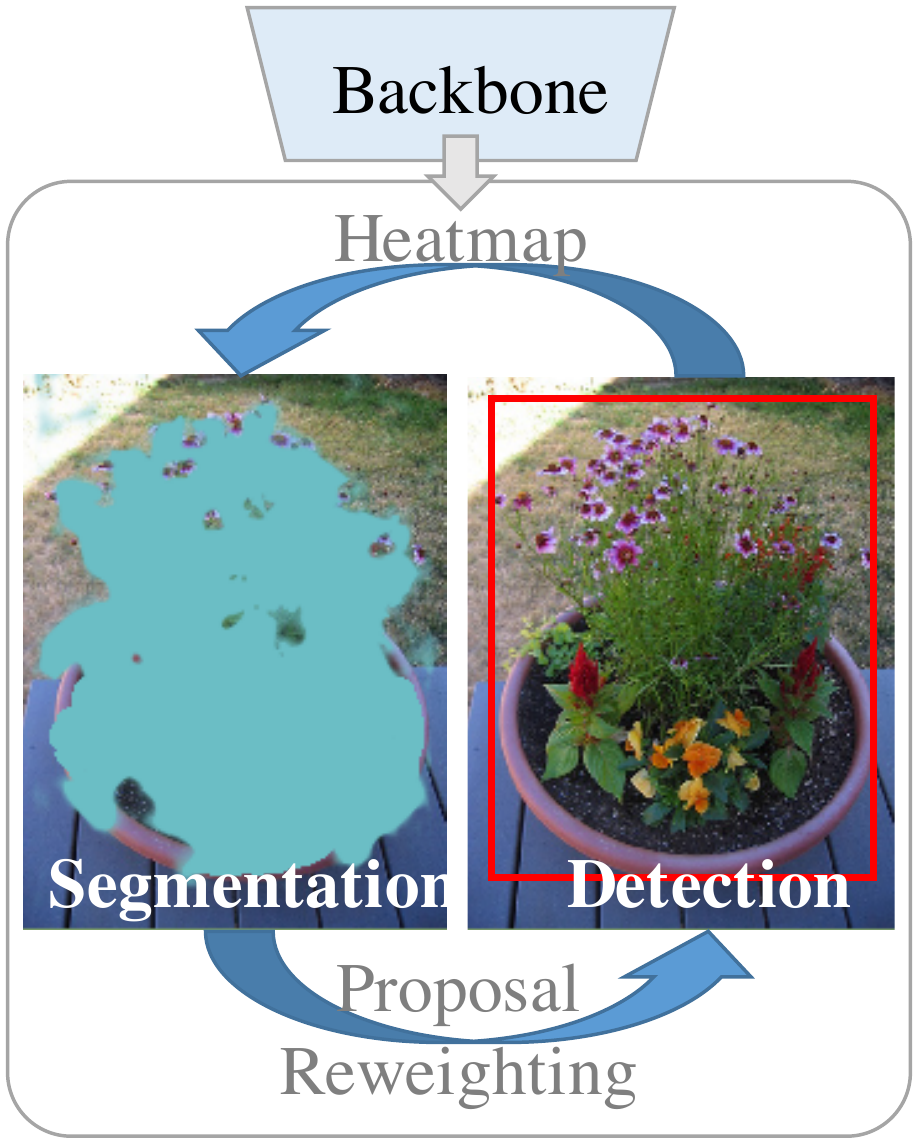}
        			\label{fig:our-pipeline}}
	\end{center}
	\caption{The schematic diagram of the previous works with segmentation utilization~\cite{diba2017weakly, wei2018ts2c} and the proposed collaboration approach. In \cite{diba2017weakly, wei2018ts2c}, a two-stage paradigm is used, in which proposals are first filtered and then detection is performed on these remaining boxes (\cite{diba2017weakly} shares the backbone between two modules). In our approach, detection and segmentation modules instruct each other in a dynamic collaboration loop in the training process.}
	\label{fig:pipeline}
	\vspace{-18pt}	
\end{figure}
  
   Recently, the most high-profile works on weakly supervised object detection all exploit the multiple instance learning (MIL) paradigm~\cite{blaschko2010simultaneous, deselaers2010localizing, siva2011weakly, siva2012defence, russakovsky2012object, gokberk2014multi, bilen2014weakly, song2014on, song2014weakly, bilen2016weakly, diba2017weakly}. Based on the assumption that the object bounding box should be the one contributing most to image classification among all proposals, the MIL based approaches work in an attention-like mechanism: automatically assign larger weights to the proposals consistent with the classification labels. 
Several promising works combining MIL with deep learning~\cite{bilen2016weakly, tang2017multiple, wei2018ts2c} have greatly pushed the boundaries of weakly supervised object detection. However, as noted in \cite{tang2017multiple, wei2018ts2c}, these methods are easy to over-fit on object parts, because the most discriminative classification evidence may derive from the entire object region, but may also from the crucial parts. The attention mechanism is effective in selecting the discriminative boxes, but does not guarantee the completeness of a detected object.  For a more reasonable inference, a further elaborative mechanism is necessary.

    Meanwhile, the completeness of a detected region is easier to ensure in weakly supervised segmentation.
    One common way to outline whole class-related segmentation regions is recurrently discovering and masking these regions in several forward passes~\cite{wei2017advers}. These segmentation maps can potentially constrain the weakly supervised object detection, given that a proposal having low intersection over union (IoU) with the corresponding segmentation map is not likely to be an object bounding box. In \cite{diba2017weakly, wei2018ts2c}, weakly supervised segmentation maps are used to filter object proposals and reduce the difficulty of detection, as shown in Fig.~\ref{fig:pre-pipeline}. However, these approaches adopt cascaded or independent models with relatively coarse segmentations to do ``hard'' delete on the proposals, inevitably resulting in a drop of the proposal recall. In a word, these methods underutilize the segmentation and limit the improvements of weakly supervised object detection.
    
    The MIL based object detection approaches and semantic segmentation approaches focus on restraining different aspects of the weakly supervised localization and have opposite strengths and shortcomings. The MIL based object detection approaches are precise in distinguishing object-related regions and irrelevant surroundings, but incline to confuse entire objects with parts due to its excessive attention to the significant regions. Meanwhile, the weakly supervised segmentation is able to cover the entire instances, but tends to mix irreverent surroundings with real objects. This complementary property is verified by Table~\ref{tab:recall}, that the segmentation can achieve a higher pixel-wise recall but lower precision, while the detection can achieve a higher pixel-level precision but lower recall. Rather than working independently, the two are naturally cooperative and can work together to overcome their intrinsic weaknesses.
    
\begin{table}[t]
\setlength{\abovecaptionskip}{-0.cm}
\setlength{\belowcaptionskip}{-0.cm}
\small{
\begin{center}
\begin{tabular}{c|c|c}
\hline  \hline 
Task                & Recall     &  Precision  \\ \hline
Weakly supervised detection   	  & 62.9\%  &  \textbf{46.3}\%      \\
Weakly supervised segmentation  & \textbf{69.7}\%  &  35.4\%    \\ \hline \hline
\end{tabular}
\end{center}
}
\caption{Pixel-wise recall and precision of detection and segmentation results on the VOC 2007 test set, following the same setting in Sec.~\ref{subsec:abl}. For a comparable pixel-level metric, the detection results are converted to the equivalent segmentation maps in a similar way described in Sec. \ref{subsec:col}. 
}
\label{tab:recall}
\vspace{-18pt}
\end{table}

    In this work, we propose a segmentation-detection collaborative network (SDCN) for more precise object detection under weak supervision, as shown in Fig.~\ref{fig:our-pipeline}. In the proposed SDCN, the detection and segmentation branches work in a collaborative manner to boost each other. Specifically, the segmentation branch is designed as a generative adversarial localization structure to sketch the object region. The detection module is optimized in an MIL manner with the obtained segmentation map serving as spatial prior probabilities of the object proposals. Besides, the object detection branch also provides supervision back to the segmentation branch by a synthetic heatmap generated from all proposal boxes and their classification scores. Therefore, these two branches tightly interact with each other and form a dynamic cooperating loop. Overall, the entire network is optimized under weak supervision of the classification loss in an end-to-end manner, which is superior to the cascaded or independent architectures in previous works~\cite{diba2017weakly, wei2018ts2c}.
 
    In summary, we make three contributions in this paper: 1) the segmentation-detection collaborative mechanism enforces deep cooperation between two complementary tasks and boosts valuable supervision to each other under the weakly supervised setting; 2) for the segmentation branch, the novel generative adversarial localization strategy enables our approach to produce more complete segmentation maps, which is crucial for improving both the segmentation and the detection branches; 3) as demonstrated in Section~\ref{sec:exp}, we achieve the best performance on PASCAL VOC 2007 and 2012 datasets, surpassing the previous state-of-the-arts.


\section{Related works}
    \textbf{Multiple Instance Learning (MIL).} MIL~\cite{dietterich1997solving} is a concept in machine learning, illustrating the essence of inexact supervision problem, in which only coarse-grained labels are available~\cite{zhou2018a}. Formally, given a training image $\mathbf{I}$, all instances in some specific form constitute a ``bag". \Eg object proposals (in detection task) or image pixels (in segmentation task) can be different forms of instances. If the image $\mathbf{I}$ is labeled with class $c$, then the ``bag" of $\mathbf{I}$ is positive with regard to $c$, meaning that there is at least one positive instance of class $c$ in this bag. If $\mathbf{I}$ is not labeled with class $c$, the corresponding ``bag" is negative to $c$ and there is no instance of class $c$ in this image. The MIL models aim at predicting the label of an input bag, and more importantly, finding positive instances in positive bags. 

\begin{figure*}[t]
	\vspace{-18pt}	
	\setlength{\abovecaptionskip}{-0.2cm}
	\setlength{\belowcaptionskip}{-0.cm}
	\begin{center}
		\includegraphics[width=0.82\linewidth]{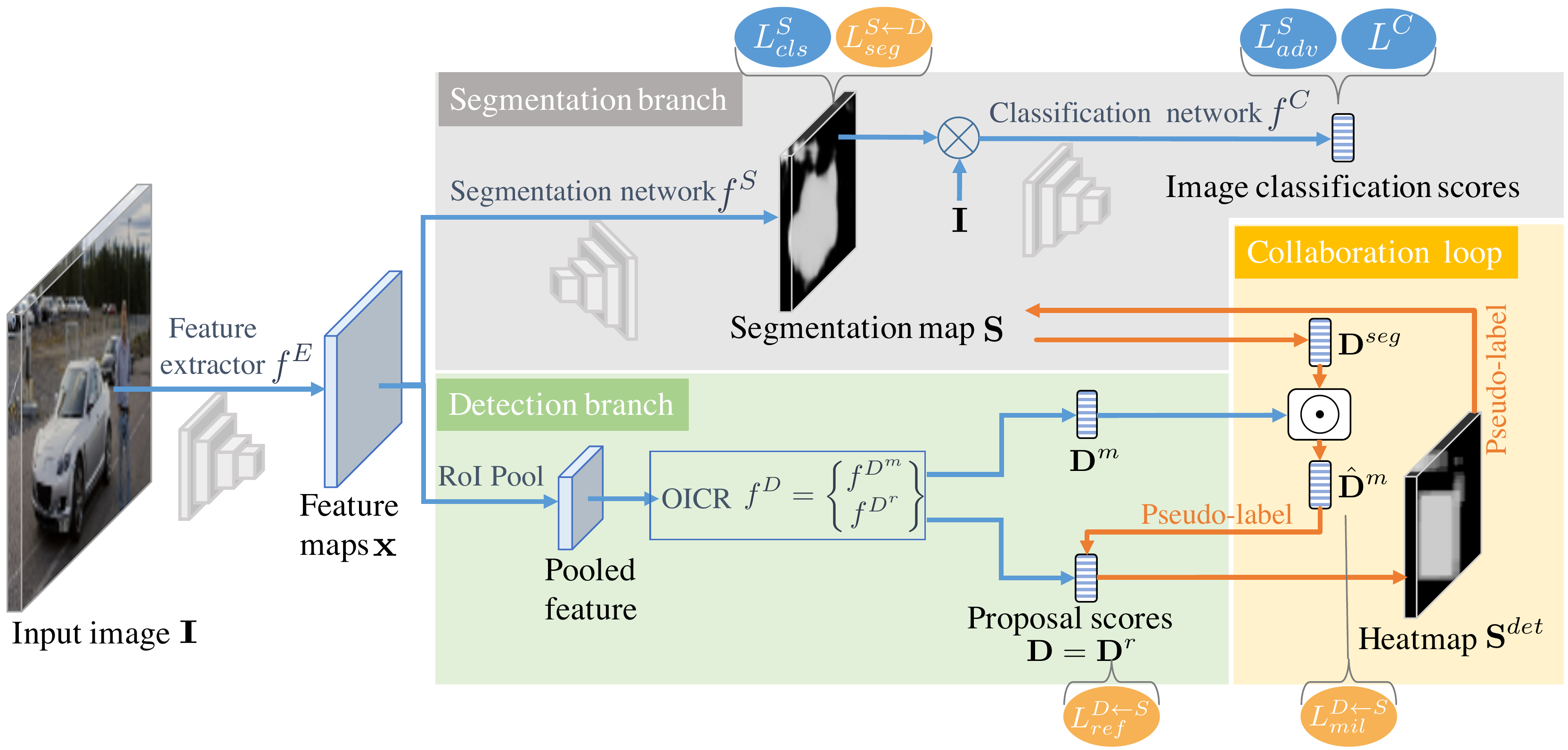}
	\end{center}
	\caption{The overall architecture. The SDCN is composed of three modules: the feature extractor, the segmentation branch, and the detection branch. The segmentation branch is instructed by a classification network in a generative adversarial learning manner, while the detection branch employs a conventional weakly supervised detector OICR~\cite{tang2017multiple}, guided by an MIL objective. These two branches further supervise each other in a collaboration loop. The solid ellipses denote the cost functions. The operations are denoted as blue arrows, while the collaboration loop is shown with orange ones. }
	\label{fig:architecture}
	\vspace{-15pt}
\end{figure*}
  
    \textbf{Weakly Supervised Object Detection.} 
    Recently, the incorporation of deep neural networks and MIL significantly improves the previous state-of-the-arts. Bilen\cite{bilen2016weakly} proposed a Weakly Supervised Deep Detection Network (WSDDN) composing of two branches acting as a proposal selector and a proposal classifier, respectively. The idea, detecting objects by the attention-based selection, is proved to be so effective that most of the latter works follow it. \Eg, WSDDN is further improved by adding recursive refinement branches in \cite{tang2017multiple}. 
    Besides these single-stage approaches, researchers have also considered the multiple-stage methods in which fully-supervised detectors are trained with the boxes detected by the single-stage methods as pseudo-labels. Zhang \cite{zhang2018zigzag} proposed a metric to estimate image difficulty with the proposal classification scores of WSDDN, and progressively trained a Fast R-CNN with curriculum learning strategy. 
To speed up the weakly supervised object detectors, Shen \cite{shen2018generative} used WSDDN  as an instructor which guides a fast generator to produce similar detection results. 
        
     \textbf{Weakly Supervised Object Segmentation.} Another  route for localizing objects is semantic segmentation. To obtain weakly supervised segmentation map, in \cite{kolesnikov2016seed}, Kolesnikov took segmentation map as an output of the network and then aggregated it to a global classification prediction to learn with category labels. In \cite{Durand2017WILDCAT}, the aggregation function is improved to incorporate both negative and positive evidence, representing both the absence and presence of the target class. 
In \cite{wei2017advers}, a recurrent adversarial erasing strategy is proposed to mask the response region of the previous forward passes and force to generate responses on other undetected parts during the current forward pass. 
          
     \textbf{Utilization of Segmentation in Weakly Supervised Detection.} Researchers have found that there are inherent relations between the weakly supervised segmentation and detection tasks. In \cite{diba2017weakly}, a segmentation branch generating coarse response maps is used to eliminate proposals unlikely to cover any objects. In \cite{wei2018ts2c}, the proposal filtering step is based on a new objectness rating TS2C defined with the weakly supervised segmentation map. 
     Ge \cite{ge2018multievidence} proposed a complex framework for both weakly supervised segmentation and detection, 
     where results from segmentation models are used as both object proposal generator and filter for the latter detection models. These methods incorporate the segmentation to overcome the limitations of weakly supervised object detection, which are reasonable and promising considering their superiorities over their baseline models. However, they ignore 
     the mentioned complementarity of these tasks and only exploit one-way cooperation, as shown in Fig.~\ref{fig:pre-pipeline}. The suboptimal manners in using the segmentation information limit the performance of their methods. 

\section{Method}
The overall architecture of the proposed segmentation-detection collaborative network (SDCN) is shown in Fig.~\ref{fig:architecture}. The network is mainly composed of three components: a backbone feature extractor $f^{E}$, a segmentation branch $f^S$, and a detection branch $f^D$. For an input image $\mathbf{I}$ 
, its feature $\mathbf{x} = f^E(\mathbf{I})$ is extracted by the extractor $f^E$, and then feeds into $f^S$ and $f^D$ for segmentation and detection, respectively. The entire network is guided by the classification labels $\mathbf{y}=[y_1,y_2,\cdots,y_N] \in \{0,1\}^N$, (where $N$ is the number of object classes), which is formatted as an adversarial classification loss and an MIL objective. Additional collaboration loss is designed for improving the accuracy of both branches in a manner of the collaborative loop. 

In \ref{subsec:det}, we first briefly introduce our detection branch, which follows the Online Instance Classifier Refinement (OICR)~\cite{tang2017multiple}. The proposed segmentation branch and collaboration mechanism are described in detail in \ref{subsec:seg} and \ref{subsec:col}.

\subsection{Detection Branch}
\label{subsec:det}
The detection branch $f^D$ aims at detecting object instances in an input image, given only image category labels. The design of $f^D$ follows the OICR~\cite{tang2017multiple}, which works in a similar fashion to the Fast RCNN~\cite{girshick2015fast}. Specifically, $f^D$ takes the feature $\mathbf{x}$ from the backbone $f^E$ and object proposals $\mathbf{B}=\{\mathbf{b}_1,\mathbf{b}_2,\dots,\mathbf{b}_B\}$ (where $B$ is the number of proposals) from Selective Search~\cite{uijlings2013selective} as input, and detects by classifying each proposal, formulated as below:
\begin{equation}
\mathbf{D} = f^D(\mathbf{x}, \mathbf{B}),\, \, \, \,  \mathbf{D} \in [0,1]^{B\times (N+1)},
\end{equation}
where $N$ denotes the number of classes with the $(N+1)^{th}$ class as the background. Each element $\mathbf{D}(i,j)$ indicates the probability of the $i^{th}$ proposal $\mathbf{b}_i$ belonging to the $j^{th}$ class.

The detection branch $f^D$ 
consists of two sub-modules, a multiple instance detection network (MIDN) $f^{D^m}$ and an online instance classifier refinement module $f^{D^r}$. The MIDN $f^{D^m}$ serves as an instructor of the refinement module $f^{D^r}$, while $f^{D^r}$ produces the final detection output.

The MIDN is the same as the mentioned WSDDN \cite{bilen2016weakly}, which computes the probability of each proposal belonging to 
each class under the supervision of category  label, with an MIL objective (in Eq. (1) of \cite{tang2017multiple}) formulated as follows:
{\setlength\abovedisplayskip{5pt}
\setlength\belowdisplayskip{5pt}
\begin{align}
\mathbf{D}^m &= f^{D^m}(\mathbf{x}, \mathbf{B}),\, \, \, \,  \mathbf{D}^m \in [0,1]^{B\times N}, \label{eq:dm}\\
L_{mil}^{D}  &= \sum\nolimits_{j=1}^N L_{BCE} \left(\sum\nolimits_{i=1}^B \mathbf{D}^m(i,j), \mathbf{y}(j)\right),
\label{eq:det_mil}
\end{align}
}where $\sum\nolimits_{i=1}^B \mathbf{D}^m(i,j)$ (denoted as $\phi_c$ in \cite{tang2017multiple}) shows the probability of an input image belonging to the $j^{th}$ category by summing up that of all proposals, and $L_{BCE}$ denotes the standard multi-class binary cross entropy loss.

Then, the resulting probability $\mathbf{D}^m$ from minimizing Eq. (\ref{eq:det_mil}) is used to generate pseudo instance classification labels for the refinement module. This process 
is denoted as:
{\setlength\abovedisplayskip{5pt}
\setlength\belowdisplayskip{5pt}
 \begin{align}
\mathbf{Y}^r=\kappa(\mathbf{D}^m),\, \, \, \, \mathbf{Y}^r \in \{0,1\}^{B\times (N+1)}.
\label{eq:ref_label}
\end{align}}Each binary element $\mathbf{Y}^r(i,j)$ indicates if the $i^{th}$ proposal is labeled as the $j^{th}$ class. 
$\kappa$ denotes the conversion from the soft probability matrix $\mathbf{D}^m$ to discrete instance labels $\mathbf{Y}^r$, where the top-scoring proposal and its highly overlapped ones are labeled as the image label and the rest are labeled as the background. Details are referred to Sec. 3.2 in \cite{tang2017multiple}.

The online instance classifier refinement module $f^{D^r}$ performs detection proposal by proposal and further constrains the spatial consistency of the detection results with the generated labels $\mathbf{Y}^r$, which is formulated as below:
{\setlength\abovedisplayskip{5pt}
\setlength\belowdisplayskip{5pt}
 \begin{align}
 \mathbf{D}^r(i,:) &= f^{D^r}(\mathbf{x}, \mathbf{b}_i), \, \, \, \,  \mathbf{D}^r \in [0,1]^{B\times (N+1)}, \\
L_{ref}^{D} &= \sum\nolimits_{j=1}^{N+1}\sum\nolimits_{i=1}^BL_{CE}\left(\mathbf{D}^r(i,j), \mathbf{Y}^r(i,j)\right),
\label{eq:det_refine}
\end{align}}where $\mathbf{D}^r(i,:)\in [0,1]^{N+1}$ is a row of $\mathbf{D}^r$, indicating the classification scores for proposal $\mathbf{b}_i$. $L_{CE}$ denotes the weighted cross entropy (CE) loss function in Eq. (4) of \cite{tang2017multiple}. Here, $L_{CE}$ is employed instead of $L_{BCE}$ considering that each proposal has one and only one positive category label. 

Eventually, the detection results are given by the refinement module, \ie $\mathbf{D} =  \mathbf{D}^r$, and the overall objective for the detection module is a combination of Eq. (\ref{eq:det_mil}) and Eq. (\ref{eq:det_refine}): 
{\setlength\abovedisplayskip{5pt}
\setlength\belowdisplayskip{5pt}
\begin{equation}
L^D =  \lambda_{mil}^{D} L_{mil}^{D} +  \lambda_{ref}^{D} L_{ref}^{D},
\label{eq:det0}
\end{equation}}where $\lambda_{mil}^{D}$ and  $\lambda_{ref}^{D}$ are balancing factors for the loss.

After optimization according to Eq. (\ref{eq:det0}), the refinement module $f^{D^r}$ can do object detection independently by discarding the MIDN in testing. 

\subsection{Segmentation Branch}
\label{subsec:seg}
Generally, the MIL weakly supervised object detection module is subject to over-fitting on discriminative parts, since smaller regions with less variation are more likely to have high consistency across the whole training set. To overcome this issue, the completeness of a detected object needs to be measured and adjusted, \eg by comparing with a segmentation map. Therefore, a weakly supervised segmentation branch is proposed to cover the complete object regions with generative adversarial localization strategy.

In detail, the segmentation branch $f^{S}$ takes the feature $\mathbf{x}$ as input and predicts a segmentation map, as below,
{\setlength\abovedisplayskip{5pt}
\setlength\belowdisplayskip{5pt}
\begin{align}
\mathbf{S} &= f^{S}(\mathbf{x}), \, \, \, \, \mathbf{S}\in [0,1]^{(N+1)\times h \times w},\\
\hspace{-0.2cm}\mathbf{s}_{k} \triangleq\mathbf{S}(k,:,: ), \,\,\,\, &k \in \{1,\dots,N+1\},  \, \, \, \, \mathbf{s}_k\in [0,1]^{h \times w}
\end{align}}where $\mathbf{S}$ has $N+1$ channels. Each channel $\mathbf{s}_{k}$ corresponds to  a segmentation map for the $k^{th}$ class with a size of $h\times w$. 

To ensure that the segmentation map $\mathbf{S} $ covers the complete object regions precisely,  a novel generative adversarial localization strategy is 
designed as adversarial training between the segmentation predictor $f^S$ and an independent image classifier $f^C$, severing as generator and discriminator respectively, as shown in Fig.~\ref{fig:architecture}. The training target of the generator $f^S$ is to fool $f^C$ into misclassifying by masking out the object regions, and the discriminator $f^C$  aims to eliminate the effect of the erased regions and correctly predict the category labels. The $f^S$ and $f^C$ are optimized alternatively, given the other one fixed. 

%

Here, we first introduce the optimization of the segmentation branch $f^S$, given the classifier $f^C$ fixed. Overall, the objective of the segmentation branch $f^S$ can be formulated as a sum of losses for each class,
\begin{equation}
L^{S} (\mathbf{S}) = L^{S}(\mathbf{s}_{1}) + L^{S}(\mathbf{s}_{2}) + \dots + L^{S}(\mathbf{s}_{N+1}).
\label{eq:seg0}
\end{equation}
Here, $L^{S}(\mathbf{s}_{k})$ is the loss for the $i^{th}$ channel of the segmentation map, consisting of an adversarial loss $L_{adv}^{S}$ and a classification loss $L_{cls}^{S}$, described in detail as following.

If the $k^{th}$ class is a \textbf{positive foreground class} \footnote{\label{foot:posneg}A positive foreground class means that the foreground class presents in the current image, while a negative one means that it does not appear.}, the segmentation map $\mathbf{s}_{k}$ should fully cover the region of the $k^{th}$ class, but does not overlap with the regions of the other classes. In other words, for an accurate $\mathbf{s}_{k}$, only the object region masked out by $\mathbf{s}_{k}$ should be classified as the $k^{th}$ class, while its complementary region should not. Formally, this expectation can be satisfied by minimizing the function
\begin{align}
\begin{split}
	L_{adv}^{S}(\mathbf{s}_k) =& L_{BCE}(f^C(\mathbf{I}*\mathbf{s}_{k}),\tilde{ \mathbf{y}}) + 
	\\& L_{BCE}(f^C(\mathbf{I}*(1-\mathbf{s}_{k})), \hat{\mathbf{y}}),
\end{split}
\label{eq:seg_adv}
\end{align}
where $*$ denotes pixel-wise product. The first term represents that the object region covered by the generated segmentation map, \ie $\mathbf{I}*\mathbf{s}_k$, should be recognized as the $k^{th}$ class by the classifier $f^C$, but does not respond to any other classes with the label $\tilde{\mathbf{y}}\in \{0,1\}^N$, where $\tilde{\mathbf{y}}(k)=1$ and $\tilde{\mathbf{y}}(i\ne k)=0$. The second term means that when the region related to the $k^{th}$ class is masked out from the input, \ie $\mathbf{I}*(1-\mathbf{s}_{k})$, the classifier $f^C$ should not recognize the $k^{th}$ class anymore without influence on the other classes, with the label $\hat{\mathbf{y}}\in \{0,1\}^N$, where $\hat{\mathbf{y}}(k)=0$ and $\hat{\mathbf{y}}(i\ne k)=\mathbf{y}(i\ne k)$.  Here, we note that generally the mask can be applied to the image $\mathbf{I}$ or the input of any layer of the classifier $f^{C}$, and since $f^C$ is fixed, the loss function in Eq. (\ref{eq:seg_adv}) only penalizes the segmentation branch $f^S$.


If the $k^{th}$ class is a \textbf{negative foreground class}
, the $\mathbf{s}_k$ should be all-zero, as no instance of this foreground class presents. This is restrained with a response constraint term. In this term, the top 20\% response pixels of each map $\mathbf{s}_k$ are pooled and averaged for a classification predication optimized with a binary cross entropy loss as below,
\begin{equation}
L_{cls}^{S}(\mathbf{s}_k)= L_{BCE}\left(\text{avgpool}_{20\%} \mathbf{s}_k, \mathbf{y}(k)\right).
\label{eq:seg_cls}
\end{equation}
If the $k^{th}$ class is labeled as negative, $\text{avgpool}_{20\%}\mathbf{s}_k$ is enforced to be close to 0, \ie all elements of the map $\mathbf{s}_k$ should approximately be 0. However, the above loss is also applicable when the $k^{th}$ class is positive, $\text{avgpool}_{20\%}\mathbf{s}_k$ should be close to 1, agreeing with the constraint in Eq. (\ref{eq:seg_adv}). 

The \textbf{background} is taken as a special case. In Eq. (\ref{eq:seg_adv}), though the labels $\tilde{\mathbf{y}}$ and $\hat{\mathbf{y}}$ do not involve the background class, the background segmentation map $\mathbf{s}_{N+1}$ is also applicable same as the other classes. When $\mathbf{s}_{N+1}$ is multiplied as the first term in Eq. (\ref{eq:seg_adv}), the target label should be all-zero $\tilde{ \mathbf{y}} =\mathbf{0}$; when $1-\mathbf{s}_{N+1}$ is used as the mask in the second term of Eq. (\ref{eq:seg_adv}), the target label should be exactly the same as the original label $\hat{\mathbf{y}}=\mathbf{y}$. For Eq. (\ref{eq:seg_cls}), we assume that a background region always appears in any input image, \ie $\mathbf{y}(N+1)=1$ for all images.

Overall, the total loss of the segmentation branch in Eq. (\ref{eq:seg0}) can be summarized and rewritten as follows,
{\setlength\abovedisplayskip{5pt}
\setlength\belowdisplayskip{5pt}
\begin{equation}
L^{S} =   \lambda_{adv}^{S} \sum_{k \text{ if }\mathbf{y}(k)=1} L_{adv}^{S}(\mathbf{s}_k) + \lambda_{cls}^{S} \sum_{k=1}^{N+1}L_{cls}^{S}(\mathbf{s}_k),
\label{eq:seg}
\end{equation}}where $\lambda_{adv}^{\ {S}}$ and  $\lambda_{cls}^{\ {S}}$ denote balance weights.

After optimizing Eq. (\ref{eq:seg}), following the adversarial manner, the segmentation branch $f^S$ is fixed, and the  classifier $f^C$ is further optimized with the following objective,
{\setlength\abovedisplayskip{13pt}
\setlength\belowdisplayskip{5pt}
\begin{align}
	L_{adv}^{C}(\mathbf{s}_k) &= L_{BCE}(f^C(\mathbf{I}*(1-\mathbf{s}_{k})), \mathbf{y}),\\
	L^C &= L_{BCE}(f^C(\mathbf{I}), \mathbf{y}) + \hspace{-0.2cm}\sum_{k \text{ if }\mathbf{y}(k)=1} L_{adv}^{C}(\mathbf{s}_k).
\label{eq:cls_adv}
\end{align}}The objective $L^C$ consists of a classification loss and an adversarial loss $L_{adv}^{C}$. The target of the classifier $f^C$ should always be $\mathbf{y}$, since it aims at digging out the remaining object regions, even if $\mathbf{s}_k$ is masked out.

Our idea for designing the segmentation branch shares the same adversarial spirit with \cite{wei2017advers}, but our design is more efficient compared with \cite{wei2017advers} that recurrently performs several forward passes for one segmentation map. Besides, we do not have the trouble of deciding number recurrent steps as \cite{wei2017advers}, which may vary with different objects.

\subsection{Collaboration Mechanism}
\label{subsec:col}

A dynamic collaboration loop is designed to complement both detection and segmentation for more accurate predictions, namely neither so large that cover the background nor so small that degenerate to object parts. 

\textbf{Segmentation instructs Detection.} 
As mentioned, the detection branch is easy to over-fit to discriminative parts, while the segmentation can cover the whole object region. So naturally, the segmentation map can be used to refine the detection results by making the proposal having a larger IoU with the corresponding segmentation map have a higher score. This is achieved by re-weighting the instance classification probability matrix $ \mathbf{D}^{m}$ in Eq. (\ref{eq:dm}) in the detection branch by using a prior probability matrix $\mathbf{D}^{seg}$ stemming from the segmentation map as follows,
{\setlength\abovedisplayskip{5pt}
\setlength\belowdisplayskip{5pt}
\begin{align}
\hat{\mathbf{D}}^m &= \mathbf{D}^{m} \odot \mathbf{D}^{seg},
\label{eq:reweight}
\end{align}}where $\mathbf{D}^{seg}(i,k)$ denotes the overlap degree between the $i^{th}$ object proposal and the connected regions from the $k^{th}$ segmentation map. $\mathbf{D}^{seg}$ is generated as below:
{\setlength\abovedisplayskip{5pt}
\setlength\belowdisplayskip{5pt}
\begin{equation}
\mathbf{D}^{seg}(i,k)= \max\nolimits_j\text{IoU}(\hat{\mathbf{s}}_{kj}, \mathbf{b}_i) + \tau_0.
\label{eq:dseg}
\end{equation}}Here, $\hat{\mathbf{s}}_{kj}$ denotes the $j^{th}$ connected component in the segmentation map $\mathbf{s}_k$, and $\text{IoU}(\hat{\mathbf{s}}_{kj}, \mathbf{b}_i)$ denotes the  intersection over union between $\hat{\mathbf{s}}_{kj}$ and the object proposal $\mathbf{b}_i$. The constant $\tau_0$ adds a fault-tolerance for the segmentation branch. 
Each column of $\mathbf{D}^{seg}$ is normalized by its maximum value, to make it range within [0, 1].

With the re-weighting in Eq. (\ref{eq:reweight}), the object proposals 
only focusing on local parts are assigned with lower weights, while those proposals precisely covering the object stand out. The connected components are employed to alleviate the issue of multiple instance occurrences, which is a hard case for weakly supervised object detection. The recent TS2C~\cite{wei2018ts2c} objectness rating designed for solving this issue is also tested in place of IoU with connected components, but no superiority shows in our case. 

The re-weighted probability matrix $\hat{\mathbf{D}}^m$ replaces $\mathbf{D}^m$ in Eq. (\ref{eq:det_mil}) and further instructs the MIDN as in Eq. (\ref{eq:mil_new}) and the refinement module as in Eq. (\ref{eq:ref_new}):
{\setlength\abovedisplayskip{5pt}
\setlength\belowdisplayskip{5pt}
\begin{align}
L_{mil}^{{D}\leftarrow{S}}  &= \sum\nolimits_{j} L_{BCE} \left(\sum\nolimits_{i}\hat{ \mathbf{D}}^m(i,j), \mathbf{y}(j)\right),
\label{eq:mil_new}\\
L_{ref}^{{D}\leftarrow{S}} &= \sum\nolimits_{j}\sum\nolimits_{i} L_{CE}\left(\mathbf{D}^r(i,j), \hat{\mathbf{Y}}^r(i,j)\right),
\label{eq:ref_new}
\end{align}}where $\hat{\mathbf{Y}}^r$ denotes the pseudo labels deriving from $\hat{\mathbf{D}}^m$ as that in Eq. (\ref{eq:ref_label}). Finally, the overall objective of the detection branch in Eq. (\ref{eq:det0}) is reformulated as below,
\begin{equation}
L^{ {D}\leftarrow {S}} = \lambda_{mil}^{{D}}L_{mil}^{{D}\leftarrow {S}} + \lambda_{ref}^{{D}}L_{ref}^{{D}\leftarrow{S}}.
\label{eq:detf}
\end{equation}

\textbf{Detection instructs Segmentation.} Though the detection boxes may not cover the whole object, they are effective for distinguishing an object from the background. To guide the segmentation branch, a detection heatmap $\mathbf{S}^{det} \in [0,1]^{(N+1)\times h \times w}$ is generated, which can be seen as an analog of the segmentation map.  Each channel $\mathbf{s}^{det}_k\triangleq\mathbf{S}^{det}(k,:,:)$ corresponds to a heatmap for the $k^{th}$ class. Specifically, for the positive class $k$, each proposal box contributes its classification score to all pixels within this proposal and thus generates the $\mathbf{s}^{det}_k$ by
{\setlength\abovedisplayskip{5pt}
\setlength\belowdisplayskip{5pt}
\begin{equation}
\mathbf{s}^{det}_k(p,q) = \sum\nolimits_{i\text{ if }(p,q) \in \mathbf{b}_i} \mathbf{D}(i,k),
\end{equation}}while the other  $\mathbf{s}^{det}_{k}$ corresponding to negative classes are set to zero. Then, $\mathbf{s}^{det}_{k}$ is normalized by its maximum response and the background heatmap $\mathbf{s}^{det}_{N+1}$ can be simply calculated as the complementary  set of the foreground, \ie
{\setlength\abovedisplayskip{5pt}
\setlength\belowdisplayskip{5pt}
\begin{equation}
\mathbf{s}^{det}_{N+1} = 1- \max\nolimits_{k\in\{1,\dots,N\}} \mathbf{s}^{det}_k.
\end{equation}}
To generate pseudo category label for each pixel, the soft segmentation map $\mathbf{S}^{det}$ is first discretized by taking the arguments of the maxima at each pixel and then the top 10\% pixels for each class are kept, while other ambiguous ones are ignored. The generated label is denoted by $\psi(\mathbf{S}^{det})$, and the instructive loss is formulated as below:
{\setlength\abovedisplayskip{5pt}
\setlength\belowdisplayskip{5pt}
\begin{equation}
L_{seg}^{S\leftarrow D}= L_{CE}(\mathbf{S}, \psi(\mathbf{S}^{det})).
\end{equation}}Therefore, the loss function of the whole segmentation branch in Eq. (\ref{eq:seg}) is now updated to
{\setlength\abovedisplayskip{5pt}
\setlength\belowdisplayskip{5pt}
\begin{equation}
L^{S\leftarrow D} =  L^{S} + \lambda_{seg}^{S} L_{seg}^{S\leftarrow D}.
\label{eq:segf}
\end{equation}}

\textbf{Overall Objective.}  With the updates in Eq. (\ref{eq:detf}) and Eq. (\ref{eq:segf}), the final objective for the entire network is
{\setlength\abovedisplayskip{5pt}
\setlength\belowdisplayskip{5pt}
\begin{align}
\text{argmin}_{f^{E},f^{S},f^{D}}L =  L^{S\leftarrow D} + L^{D\leftarrow S}.
\end{align}}Briefly, the above objective is optimized in an end-to-end manner. The image classifier $f^{C}$ is optimized with the loss $L^C$ alternatively, as most adversarial methods. 
The optimization can be easily conducted using gradient descent. For clarity, the training and the testing of our SDCN are summarized in Algorithm~\ref{algo:SCDN}.

In the testing stage, as shown in Algorithm~\ref{algo:SCDN}, only the  feature extractor $f^E$ and the refinement module $f^{D^r}$ 
are needed, which make our method as efficient as \cite{tang2017multiple}.

\renewcommand{\algorithmicrequire}{\textbf{Input:}}
\renewcommand{\algorithmicensure}{\textbf{Output:}}

\setlength{\textfloatsep}{8pt}
\begin{algorithm}
\caption{Training and Testing SDCN}\label{algo:SCDN}
\begin{algorithmic}[1]
\Require training set with category labels $\mathbf{T}_1 = \{(\mathbf{I}, \mathbf{y})\}$.
\Procedure{Training}{}
\State forward SDCN $f^E(\mathbf{I})\hspace{-3pt} \rightarrow \hspace{-3pt} \mathbf{x} $, $f^D(\mathbf{x})\hspace{-3pt} \rightarrow \hspace{-3pt}\mathbf{D} $, $f^S(\mathbf{x})\hspace{-3pt} \rightarrow\hspace{-3pt} \mathbf{S}$,
\State forward the classifier $f^{C}(\mathbf{s}_k*\mathbf{I})$ and $f^C((1-\mathbf{s}_k)*\mathbf{I})$,
\State generate variables $\mathbf{D}^{seg}$ and $\mathbf{S}^{det}$ with $\mathbf{S}$ and $\mathbf{D}$,
\State compute $L^{D\leftarrow S}$ in Eq.(\ref{eq:detf}) and $L^{S\leftarrow D}$ in Eq.(\ref{eq:segf}),
\State backward the loss $L=L^{D\leftarrow S}+L^{S\leftarrow D}$ for SDCN,
\State compute and backward the loss $L^C$ for $f^{C}$,
\State continue until convergence.
\EndProcedure
\Ensure the optimized SDCN ($f^{E}$ and $f^{D}$) for detection.
\end{algorithmic}
\begin{algorithmic}[1]
\Require test set $\mathbf{T}_2 = \{\mathbf{I}\}$.
\Procedure{Testing}{}
\State forward SDCN $f^{E}(\mathbf{I}) \rightarrow \mathbf{x} $, $f^{D^r}(\mathbf{x}) \rightarrow \mathbf{D} $,
\State post-process for detected bounding boxes with $\mathbf{D}$.
\EndProcedure
\Ensure the detected object bounding boxes for $\mathbf{T}_2$.
\end{algorithmic}
\end{algorithm}

\section{Experiments}
\label{sec:exp}
We evaluate the proposed segmentation-detection collaborative network (SDCN) for weakly supervised object detection to prove its advantages over the state-of-the-arts.

\begin{figure*}[t]
	\setlength{\abovecaptionskip}{-0.3cm}
	\setlength{\belowcaptionskip}{-0.cm}
	\vspace{-18pt}
	\begin{center}
		\subfloat[][Segmentation]{
        			\includegraphics[height=2.3in]{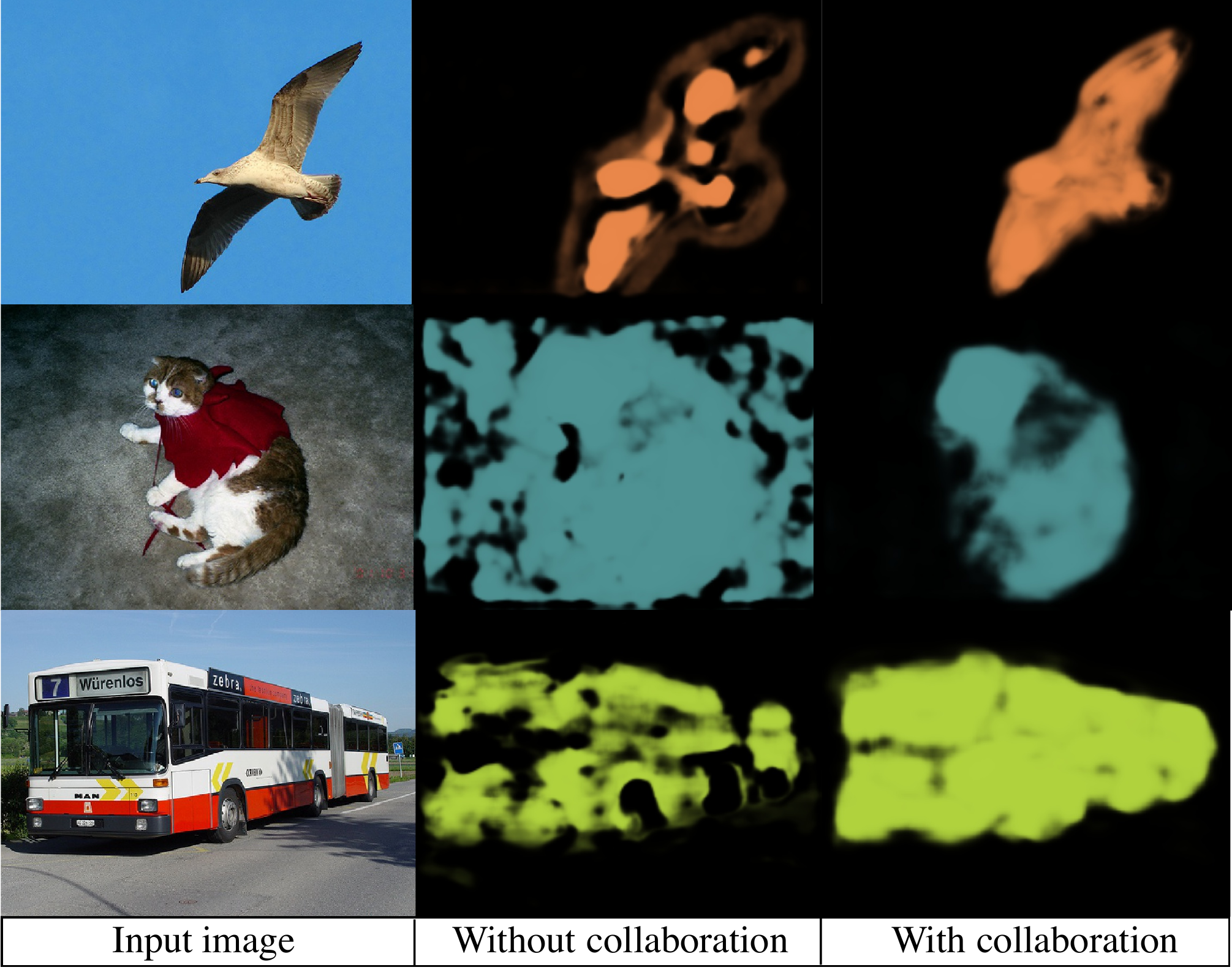}
        			\label{fig:vis_seg}}
		\subfloat[][Detection]{
        			\includegraphics[height=2.3in]{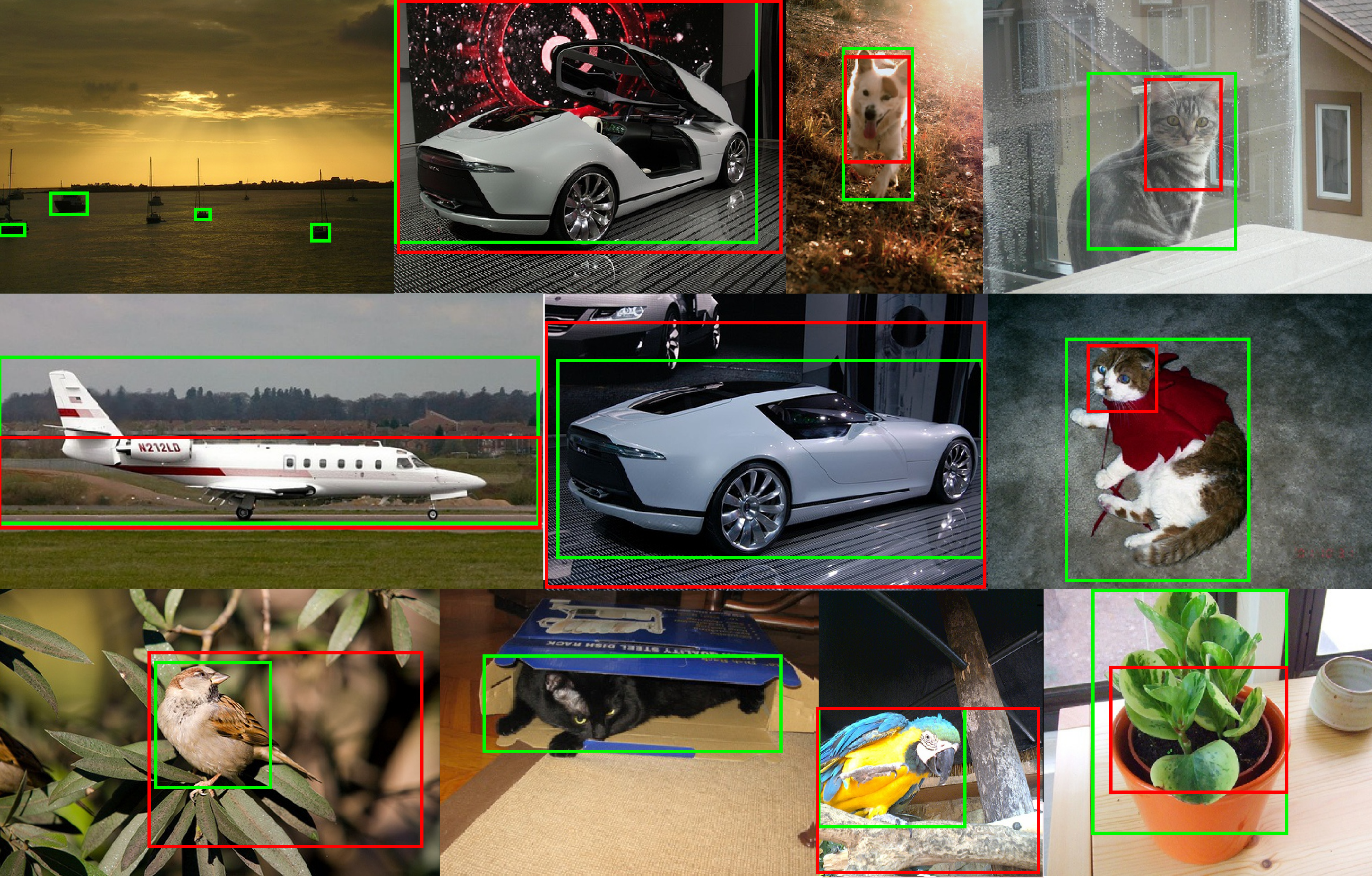}
        			\label{fig:vis_det}}
	\end{center}
	\caption{Visualization of the segmentation and the detection results without and with collaboration. In (a), the columns from left to right are the original images, the segmentation map obtained without and with the collaboration loop. In (b), the detection results of OICR\cite{tang2017multiple} without consideration of collaboration, and the proposed method with collaboration loop are shown with red and green boxes, respectively. (Absence of boxes means no detected object given the detection threshold.)}
	\label{fig:vis}
	\vspace{-15pt}
\end{figure*}

\subsection{Experimental Setup}
\textbf{Datasets.} The evaluation is conducted on two commonly used datasets for weakly supervised detection, including the PASCAL VOC 2007~\cite{everingham2010pascal} and 2012~\cite{everingham2015pascal}. The VOC 2007 dataset includes 9,963 images with total 24,640 objects in 20 classes. It is divided into a \emph{trainval} set with 5,011 images and a \emph{test} set with 4,952 images. The more challenging VOC 2012 dataset consists of 11,540 images with 27,450 objects in \emph{trainval} set and 10,991 images for test. In our experiments, the \emph{trainval} split is used for training and the \emph{test} set is for testing. The performance is reported in terms of two metrics:  1) correct localization (CorLoc)~\cite{deselaers2012weakly} on the \emph{trainval} spilt and 2) average precision (AP) on the \emph{test} set.  

\textbf{Implementation.}
For the backbone network $f^{E}$, we use the VGG-16~\cite{simonyan14very}. For $f^{D}$, the same architecture as that in OICR~\cite{wei2018ts2c} is employed. For $f^{S}$, similar segmentation header to the CPN~\cite{Chen2018CPN} is adopted. For the adversarial classifier $f^{C}$, ResNet-101~\cite{he2015deep} is used and the segmentation masking operation is applied after the res4b22 layer. The detailed architecture is shown in the Appendix~\ref{sec:arch}.

We follow a three-step training strategy: 1) the classifier $f^{C}$ is trained with a fixed learning rate $5\times10^{-4}$ until its convergence; 2) the segmentation branch $f^{S}$ and detection branch $f^D$ are pre-trained without collaboration; 3) the entire architecture is trained following the end-to-end manner. The SDCN runs for 40k iterations with learning rate $10^{-3}$, following 30k iterations with learning rate $10^{-4}$. The same multi-scale training and testing strategies in OICR~\cite{tang2017multiple} are adopted. 
To achieve balanced impacts between detection and segmentation branches, the weights of the losses are simply set to make the gradients have similar scales, i.e. $\lambda_{adv}^{S}=1$, $\lambda_{cls}^{S}=0.1$, $\lambda_{seg}^{S}=0.1$, $\lambda_{mil}^{D}=1$ and $\lambda_{ref}^{D}=1$, respectively. The constant $\tau_0$ in Eq. (\ref{eq:dseg}) is empirically set to 0.5.

\begin{table}[b]
\small{
\setlength{\abovecaptionskip}{0.cm}
\setlength{\belowcaptionskip}{-0.cm}
\begin{center}
\resizebox{\linewidth}{!}{
\begin{tabular}{c|c|c|c|c}
\hline \hline 
Det. branch	&Seg. branch	& Seg. $\rightarrow$ Det. & Det. $\rightarrow$ Seg.   & mAP  \\ \hline
$\surd$		&			&   				 & 				   & 41.2  \\ 
$\surd$		&$\surd$		& 				 & 				   & 41.3  \\ 
$\surd$		&$\surd$		& $\surd$			 & 				   & 36.8  \\ 
$\surd$		&$\surd$		& $\surd$			 & $\surd$			   &\textbf{48.3}    \\ \hline \hline
\end{tabular}
}
\end{center}
\caption{mAP (in \%) of different weakly supervised strategies with the same backbone on the VOC 2007 dataset.}
\label{tab:ablation}
}
\end{table}

\subsection{ Ablation Studies}
\label{subsec:abl}
Our ablation study is conducted on VOC 2007 dataset. 
Four weakly supervised strategies are compared and the results are shown in Table~\ref{tab:ablation}. The baseline detection method without the segmentation branch is the same as the  OICR\cite{wei2018ts2c}. Another naive consideration is directly including the detection and segmentation modules in a multi-task manner without any collaboration between them. The model where only segmentation branch instructs detection branch is also tested. Its mAP is the lowest, since the mean intersection over union (mIoU) between the segmentation results and the ground-truth drops from 37\% to 25.1\% without the guidance of detection branch, which proves that these two branches should not collaborate in one-way. Our method with segmentation-detection collaboration achieves the highest mAP. It can be observed that the proposed method improves all baseline models by large margins, demonstrating the effectiveness and necessity of the collaboration loop between detection and segmentation.

\setlength{\tabcolsep}{3pt}
\begin{table*}[h]
\vspace{-5pt}
\setlength{\abovecaptionskip}{0.2cm}
\setlength{\belowcaptionskip}{-0.cm}
\resizebox{\textwidth}{!}{
\begin{tabular}{l|cccccccccccccccccccc|r}
\hline \hline
Methods & aero & bike & bird & boat & bottle & bus &car &cat & {chair} &{ cow} & {table} & {dog} & {horse} & {mbike} & {person} &{plant} & {sheep} & {sofa} & {train} & {tv} & mAP \\ \hline\hline
 \multicolumn{22}{c}{Single-stage} \\ \hline
WSDDN-VGG16~\cite{bilen2016weakly} &39.4 &50.1 &31.5 &16.3 &12.6 &64.5 &42.8 &42.6 &10.1 &35.7 &24.9 &38.2 &34.4 &55.6 &9.4 &14.7 &30.2 &40.7 &54.7 &46.9 &34.8 \\
OICR-VGG 16~\cite{tang2017multiple} &58.0 &62.4 &31.1 &19.4 &13.0 &65.1 &62.2 &28.4 &\textbf{24.8} &44.7 &30.6 &25.3 &37.8 &65.5 &15.7 &24.1 &41.7 &46.9 &\textbf{64.3} &62.6 &41.2 \\ 
MELM-L+RL\cite{wan2018min} &50.4 &57.6 &37.7 &23.2 &13.9 &60.2 &63.1 &44.4 &24.3 &\textbf{52.0} &42.3 &42.7 &43.7 &66.6 &2.9 &21.4 &45.1 &45.2 &59.1 &56.2 &42.6 \\ 
TS2C~ \cite{wei2018ts2c} &59.3 &57.5 &\textbf{43.7} &27.3 &13.5 &63.9 &61.7 &59.9 &24.1 &46.9 &36.7 &45.6 &39.9 &62.6 &10.3 &23.6 &41.7 &52.4 &58.7 &56.6 &44.3 \\ 
\cite{tang2018weakly}       &57.9 &\textbf{70.5} &37.8 &5.7 &21.0 &66.1 &\textbf{69.2} &59.4 &3.4 &57.1 &\textbf{57.3} &35.2 &\textbf{64.2} &\textbf{68.6} &\textbf{32.8} &\textbf{28.6} &\textbf{50.8} &49.5 &41.1 &30.0 &45.3 \\ \hline
SDCN (\textbf{ours})       &\textbf{59.8} &67.1 &32.0 &\textbf{34.7} &\textbf{22.8} &\textbf{67.1} &63.8 &\textbf{67.9} &22.5 &48.9 &47.8 &\textbf{60.5} &51.7 &65.2 &11.8 &20.6 &42.1 &\textbf{54.7} &60.8 &\textbf{64.3} &\textbf{48.3}     \\ \hline 
\multicolumn{22}{c}{Multiple-stage} \\ \hline
WSDDN-Ens.~\cite{bilen2016weakly} &46.4 &58.3 &35.5 &25.9 &14.0 &66.7 &53.0 &39.2 &8.9 &41.8 &26.6 &38.6 &44.7 &59.0 &10.8 &17.3 &40.7 &49.6 &56.9 &50.8 &39.3 \\ 
HCP+DSD+OSSH3\cite{jie2017deep} &52.2 &47.1 &35.0 &26.7 &15.4 &61.3 &66.0 &54.3 &3.0 &53.6 &24.7 &43.6 &48.4 &65.8 &6.6 &18.8 &51.9 &43.6 &53.6 &62.4 &41.7\\
OICR-Ens.+FRCNN\cite{tang2017multiple} &\textbf{65.5} &67.2 &47.2 &21.6 &22.1 &68.0 &68.5 &35.9 &5.7 &63.1 &49.5 &30.3 &64.7 &66.1 &13.0 &25.6 &50.0 &\textbf{57.1} &60.2 &59.0 &47.0  \\ 
MELM-L2+ARL\cite{wan2018min} &55.6 &66.9 &34.2 &29.1 &16.4 &68.8 &68.1 &43.0 &\textbf{25.0} &65.6 &45.3 &53.2 &49.6 &68.6 &2.0 &25.4 &52.5 &56.8 &62.1 &57.1 &47.3  \\ 
ZLDN-L\cite{zhang2018zigzag}  &55.4 &68.5 &\textbf{50.1} &16.8 &20.8 &62.7 &66.8 &56.5 &2.1 &57.8 &47.5 &40.1 &69.7 &68.2 &21.6 &\textbf{27.2} &53.4 &56.1 &52.5 &58.2 &47.6 \\ 
TS2C+FRCNN~ \cite{wei2018ts2c}   &--      &--      &--      &--      &--        &--     &--     &--     &--       &--     &--       &--     &--       &--       &--        &--       &--       &--       &--       &--       & 48.0   \\ 
Ens.+FRCNN\cite{tang2018weakly} &63.0 &69.7 &40.8 &11.6 &\textbf{27.7} &\textbf{70.5} &\textbf{74.1} &58.5 &10.0 &\textbf{66.7} &\textbf{60.6} &34.7 &\textbf{75.7} &\textbf{70.3} &\textbf{25.7} &26.5 &\textbf{55.4} &56.4 &55.5 &54.9 &50.4 \\ \hline
SDCN+FRCNN (\textbf{ours})    & 61.1 &\textbf{70.6}  &40.2  &\textbf{32.8} &23.9 &63.4  &68.9  &\textbf{68.2}  &18.3  &60.2  &53.5  &\textbf{63.6}  &53.6  &66.1  &14.6 &21.8  &50.5 &56.9  &\textbf{62.4}  &\textbf{67.9}  &\textbf{51.0}    \\ \hline \hline
\end{tabular}
}
\caption{Average precision (in \%) for our method and the state-of-the-arts on VOC 2007 \emph{test} split.}
\label{tab:ap_voc2007}
\vspace{-5pt}
\end{table*}

\setlength{\tabcolsep}{3pt}
\begin{table*}[h]
\setlength{\abovecaptionskip}{0.2cm}
\setlength{\belowcaptionskip}{-0.cm}
\resizebox{\textwidth}{!}{
\begin{tabular}{l|cccccccccccccccccccc|c}
\hline \hline
Methods & aero & bike & bird & boat & bottle & bus &car &cat & {chair} &{ cow} & {table} & {dog} & {horse} & {mbike} & {person} &{plant} & {sheep} & {sofa} & {train} & {tv} & CorLoc \\ \hline\hline
\multicolumn{22}{c}{Single-stage} \\ \hline
WSDDN-VGG16~\cite{bilen2016weakly} &65.1 &58.8 &58.5 &33.1 &39.8 &68.3 &60.2 &59.6 &34.8 &64.5 &30.5 &43.0 &56.8 &82.4 &\textbf{25.5} &41.6 &61.5 &55.9 &65.9 &63.7 &53.5 \\ 
OICR-VGG16~\cite{tang2017multiple} &81.7 &80.4 &48.7 &49.5 &32.8 &\textbf{81.7} &\textbf{85.4} &40.1 &40.6 &\textbf{79.5} &35.7 &33.7 &60.5 &\textbf{88.8} &21.8 &\textbf{57.9} &\textbf{76.3} &59.9 &75.3 &\textbf{81.4} &60.6 \\ 
TS2C~\cite{wei2018ts2c} &84.2 &74.1 &\textbf{61.3} &52.1 &32.1 &76.7 &82.9 &66.6 &\textbf{42.3} &70.6 &39.5 &57.0 &61.2 &88.4 &9.3 &54.6 &72.2 &60.0 &65.0 &70.3 &61.0 \\ 
\cite{tang2018weakly} &77.5 &81.2 &55.3 &19.7 &44.3 &80.2 &86.6 &69.5 &10.1 &87.7 &68.4 &52.1 &84.4 &91.6 &57.4 &63.4 &77.3 &58.1 &57.0 &53.8 &63.8 \\ \hline
SDCN (\textbf{ours}) &\textbf{85.8} &\textbf{83.1} & 56.2  &\textbf{58.5}  &\textbf{44.7} &80.2 &85.0 &\textbf{77.9} &29.6 &78.8  &\textbf{53.6} &\textbf{74.2} &\textbf{73.1} &88.4 &18.2 &57.5 &74.2  &\textbf{60.8} &\textbf{76.1}  &79.2  &\textbf{66.8}    \\ \hline
\multicolumn{22}{c}{Multiple-stage} \\ \hline
HCP+DSD+OSSH3\cite{jie2017deep} &72.7 &55.3 &53.0 &27.8 &35.2 &68.6 &81.9 &60.7 &11.6 &71.6 &29.7 &54.3 &64.3 &88.2 &22.2 &53.7 &72.2 &52.6 &68.9 &75.5 &56.1 \\
WSDDN-Ens.~\cite{bilen2016weakly}  &68.9 &68.7 &\textbf{65.2} &42.5 &40.6 &72.6 &75.2 &53.7 &29.7 &68.1 &33.5 &45.6 &65.9 &86.1 &27.5 &44.9 &76.0 &62.4 &66.3 &66.8 &58.0 \\ 
MELM-L2+ARL\cite{wan2018min}   &--      &--      &--      &--      &--        &--     &--     &--     &--       &--     &--       &--     &--       &--       &--        &--       &--       &--       &--       &--       & 61.4  \\ 
ZLDN-L\cite{zhang2018zigzag}  &74.0 &77.8 &\textbf{65.2} &37.0 &46.7 &75.8 &83.7 &58.8 &17.5 & 73.1 & 49.0 & 51.3 & 76.7 &87.4 & \textbf{30.6}  &47.8 &75.0 &62.5 &64.8 &68.8 &61.2 \\
OICR-Ens.+FRCNN\cite{tang2017multiple}  &85.8 &82.7 &62.8 &45.2 &43.5 &\textbf{84.8} &\textbf{87.0} &46.8 &15.7 &\textbf{82.2} &51.0 &45.6 &\textbf{83.7} &\textbf{91.2} &22.2 &\textbf{59.7} &75.3 &65.1 &76.8 &78.1 &64.3 \\ 
Ens.+FRCNN\cite{tang2018weakly} &83.8 &82.7 &60.7 &35.1 &53.8 &82.7 &88.6 &67.4 &22.0 &86.3 &68.8 &50.9 &90.8 &93.6 &44.0 &61.2 &82.5 &65.9 &71.1 &76.7 &68.4\\  \hline
SDCN+FRCNN (\textbf{ours})  & \textbf{88.3} & \textbf{84.3} &59.2 &\textbf{58.5} &\textbf{47.7} & 81.2 &86.7 &\textbf{78.8}  &\textbf{29.9}  &81.5  &\textbf{54.0}  & \textbf{78.4}   &75.2   & 90.8  & 20.2   &55.3  &\textbf{76.3}  & \textbf{68.6}  & \textbf{79.1} & \textbf{82.8} &\textbf{68.8}    \\ \hline  \hline
\end{tabular}
}
\caption{CorLoc (in \%) for our method and the state-of-the-arts on VOC 2007 \emph{trainval} split.}
\label{tab:corloc_voc2007}
\vspace{-10pt}
\end{table*}

The segmentation masks and detection results without and with the collaboration are visualized in Fig.~\ref{fig:vis}. As observed in Fig.~\ref{fig:vis_seg}, with the instruction from the detection branch, the segmentation map becomes much more precise with fewer confusions between the background and the class-related region. Similarly, as shown in Fig.~\ref{fig:vis_det}, the baseline approach inclines to mix discriminative parts with target object bounding boxes, while with the guidance from segmentation the more complete objects are detected. The visualization clearly illustrates the benefits to each other.

For the validation of hyper-parameters and detailed error analysis, please refer to the Appendix~\ref{sec:abl}.

\subsection{Comparisons with state-of-the-arts}
All comparison methods are first evaluated on VOC 2007 as shown in Table~\ref{tab:ap_voc2007} and Table~\ref{tab:corloc_voc2007} in terms of mAP and CorLoc. Among single-stage methods, our method outperforms others on the most categories, leading to a notable improvement on average. Especially, our method performs much better than the state-of-the-arts on ``boat'', ``cat", ``dog", as our approach leans to detect more complete objects, though in most cases instances of these categories can be identified by parts. 
Moreover, our method produces significant improvements compared with the OICR\cite{tang2017multiple} with exactly the same architecture. The most competitive method \cite{tang2018weakly} is designed for weakly supervised object proposal, which is not really competing but complementary to our method, and replacing the fixed object proposal in our method with \cite{tang2018weakly} potentially improves the performance. 
Besides, the performance of our single-stage method is even comparable with the multiple-stage methods~\cite{tang2017multiple, wei2018ts2c,zhang2018zigzag,wan2018min}, illustrating the effectiveness of the proposed dynamic collaboration loop.

Furthermore, all methods can be enhanced by training with multiple stages, as shown at the bottom of Table~\ref{tab:ap_voc2007}. Following \cite{tang2017multiple,wei2018ts2c}, the top scoring detection bounding boxes from SDCN is used as the labels for training a Fast RCNN~\cite{girshick2015fast} with the backbone of VGG16, denoted as SDCN+FRCNN. By this simple multi-stage training strategy, the performance can be further boosted to 51\%, which surpasses all the state-of-the-art multiple-stage methods, though \cite{tang2017multiple, tang2018weakly} use more complex ensemble models. It is noted that the approaches, \eg. HCP+DSD+OSSH3\cite{jie2017deep} and ZLDN-L\cite{zhang2018zigzag}, attempt to design more elaborate training mechanism by using self-paced or curriculum learning. We believe that the performance of our model SDCN+FRCNN can be further improved by adopting such algorithms. 

\setlength{\tabcolsep}{9pt}
\begin{table}[t]
\setlength{\abovecaptionskip}{-0.2cm}
\setlength{\belowcaptionskip}{-0.cm}
\begin{center}
\resizebox{\linewidth}{!}{
\begin{tabular}{l|l|c|c}
\hline\hline
\multicolumn{2}{c|}{Methods} & mAP  & CorLoc\\ \hline \hline
\multirow{4}{*}{Single-stage}  &OICR-VGG16~\cite{tang2017multiple}   &37.9  &62.1\\ 
					     &TS2C~\cite{wei2018ts2c}                &40.0 & 64.4 \\ 
					     &\cite{tang2018weakly}			     &40.8 &64.9\\\cline{2-4}
					     &SDCN (\textbf{ours})  & \textbf{43.5} &\textbf{67.9}\\ \hline
\multirow{6}{*}{Multiple-stage}  &MELM-L2+ARL\cite{wan2018min}   & 42.4 &--  \\ 
					    &OICR-Ens.+FRCNN~\cite{tang2017multiple}   &42.5  &65.6\\ 
					    &ZLDN-L\cite{zhang2018zigzag}  &42.9 &61.5 \\ 
					    &TS2C+FRCNN~\cite{wei2018ts2c}  &44.4 &--  \\ 
					    &Ens.+FRCNN\cite{tang2018weakly} &45.7 &69.3\\ \cline{2-4}
					    &SDCN+FRCNN (\textbf{ours})  & \textbf{46.7} &\textbf{69.5}\\ \hline\hline
\end{tabular}
}
\end{center}
\caption{mAP and CorLoc (in \%) for our method and the state-of-the-arts on VOC 2012 \emph{trainval} split.}
\label{tab:voc2012}
\vspace{-5pt}
\end{table}

The comparison methods are further evaluated on the more challenging VOC 2012 dataset, as shown in Table~\ref{tab:voc2012}.  As expected, the proposed method achieves significant improvements with the same architecture as \cite{tang2017multiple,wei2018ts2c}, demonstrating its superiority again.

Overall, our SDCN significantly improves the performance of weakly supervised object detection on average, benefitting from the deep collaboration of segmentation and detection. However, there are still several classes on which the performance is hardly improved as shown in Table~\ref{tab:ap_voc2007}, \eg ``chair" and ``person". The main reason is the large portion of occluded and overlapped samples for these classes
, which leads to incomplete or connected responses on the segmentation map and bad interaction with the detection branch, leaving room for further improvements.

\textbf{Time cost.} Our training speed is roughly 2$\times$ slower than that of the baseline OICR~\cite{tang2017multiple}, but the testing time costs of our method and OICR are the same, since they share exactly the same architecture of the detection branch.

\section{Conclusions and Future Work}

In this paper, we present a novel segmentation-detection collaborative network (SDCN) for weakly supervised object detection. Different from the previous works, our method exploits a collaboration loop between segmentation task and detection task to combine the merits of both. Extensive experimental results safely reach the conclusion that our method successfully exceeds the previous state-of-the-arts, while it keeps efficiency in the inference stage. The design of SDCN may be more elaborate for densely overlapped or partially occluded objects
, which is more challenging and left as future work.

\clearpage

\appendix

\section{Appendix: Network Architecture}
\label{sec:arch}

\begin{figure*}[hb]
	\hspace{1.7cm}
	\begin{minipage}{0.35\linewidth}
		\subfloat[][Feature extractor $f^E$]{
        			\includegraphics[height=13cm]{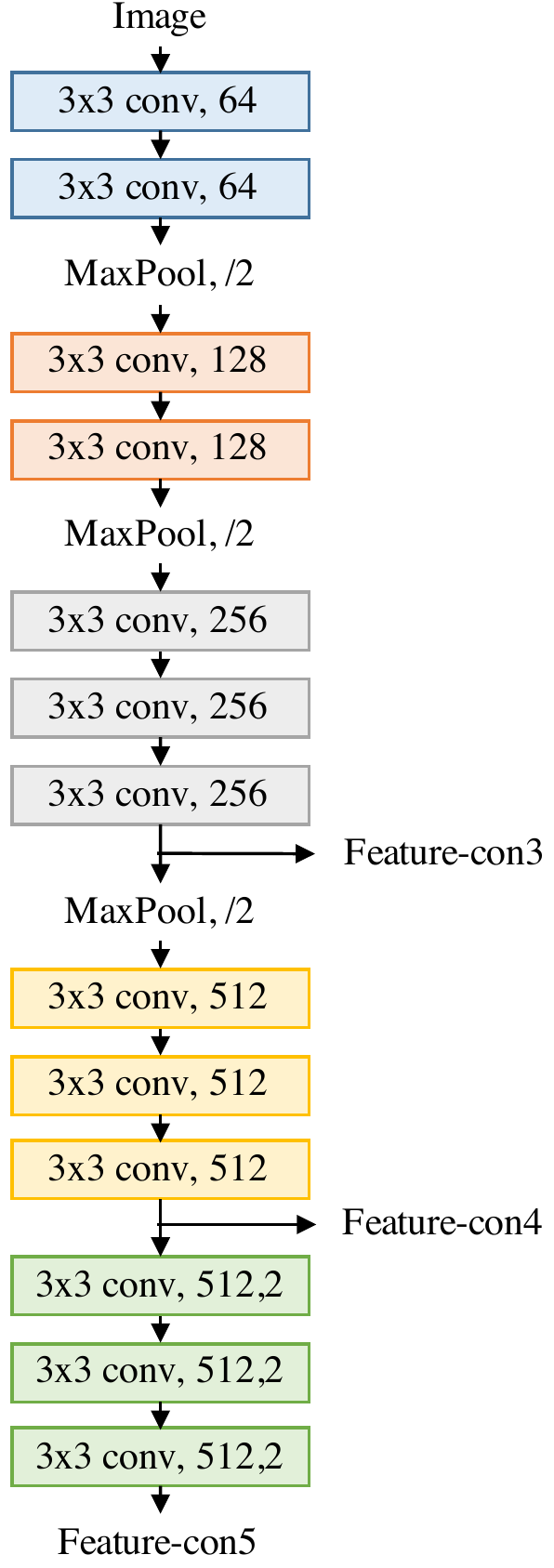}
        			\label{fig:feat}}
	\end{minipage}
	\begin{minipage}{0.65\linewidth}
		\subfloat[][Detection branch $f^D$]{
        			\includegraphics[width=8.5cm]{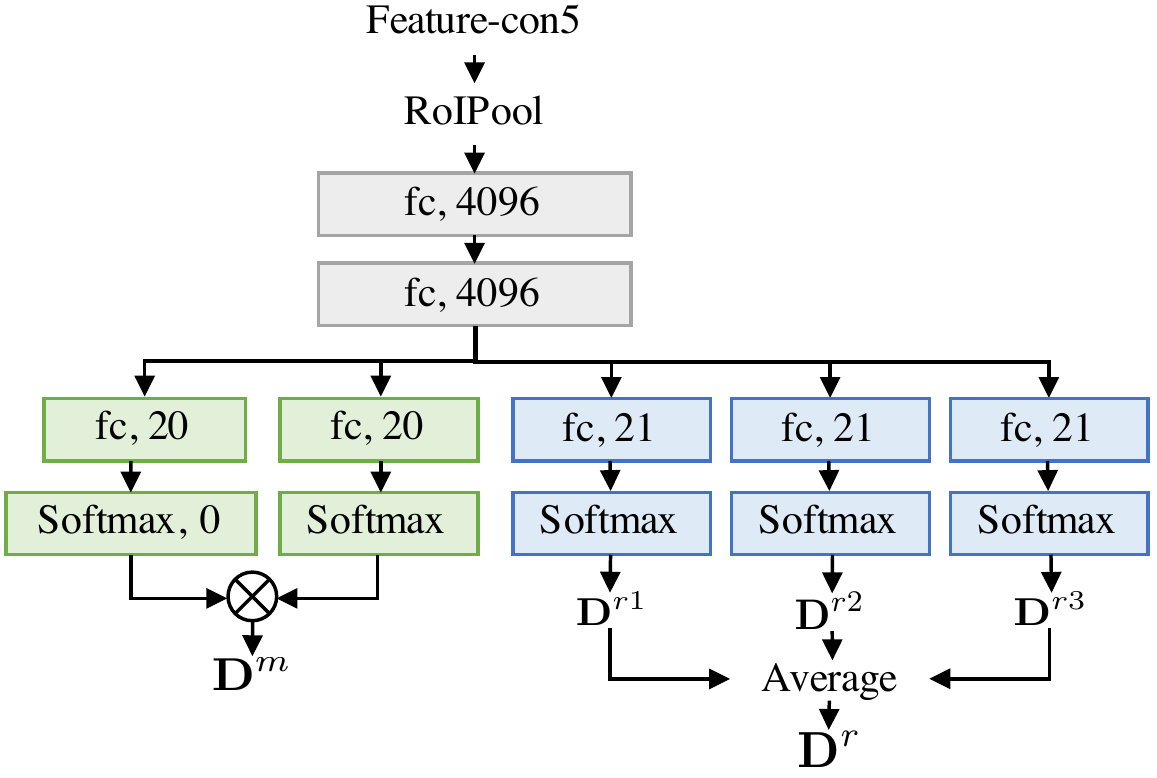}
        			\label{fig:det}}\\
		\subfloat[][Segmentation branch $f^S$]{
        			\includegraphics[width=8.5cm]{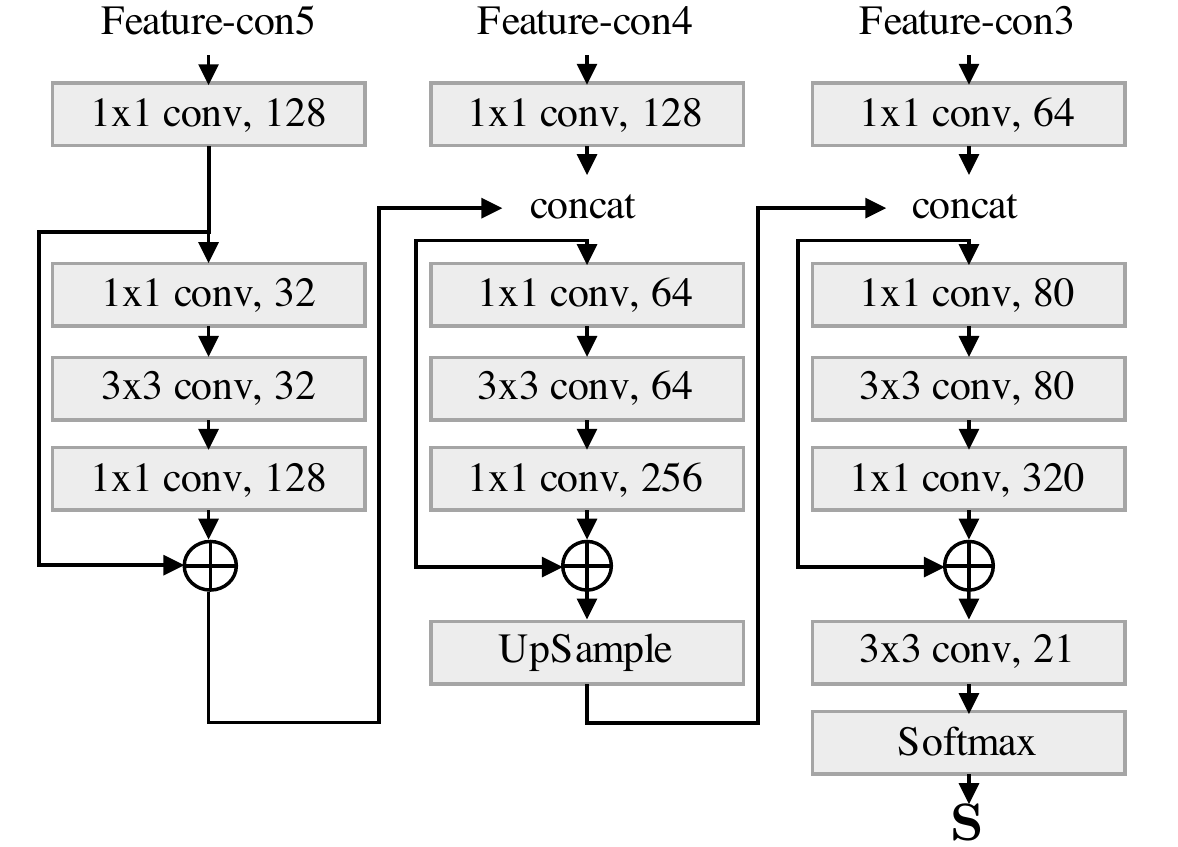}
        			\label{fig:seg}}
	\end{minipage}
	\caption{Network architectures for (a) the feature extractor, (b) the detection branch, and (c) the segmentation branch.}
	\label{fig:arch}
\end{figure*}

The network architectures of the proposed method are shown in Fig.~\ref{fig:arch}. The feature extractor $f^E$ and the detection branch $f^D$ are exactly the same as the OICR~\cite{tang2017multiple}, while the segmentation branch $f^S$ follows the design of the RefineNet in CPN~\cite{Chen2018CPN}. The classification network $f^C$ for generative adversarial localization is omitted, considering that it has exactly the same architecture with the well-known ResNet~\cite{he2015deep}.

The feature extractor $f^E$ in Fig.~\ref{fig:feat} is basically the VGG16~\cite{simonyan14very} network. The max-pooling layer after ``con4'' and its subsequent convolutional layers are replaced by the dilated convolutional layers in order to increase the resolution of the last output feature map.

The detection branch $f^D$ is composed of a multiple instance detection network (MIDN) $f^{D^m}$ and an online instance classifier refinement module $f^{D^r}$, which are shown in green and blue in Fig.~\ref{fig:det}, respectively. In MIDN, two branches are in charge of computing the instance classification weights for each proposal and classifying each proposal respectively, by performing softmax along different dimensions. For the refinement module, although the instance classifier is refined only one time for a clear illustration in the manuscript, in fact, it can be refined multiple times. We follow the OICR~\cite{tang2017multiple}, which performs the refinement 3 times, as shown in Fig.~\ref{fig:det}. The $k^{th} (k=1,2,3)$ refinement is instructed by the $(k-1)^{th}$ detection results (with the $\mathbf{D}^m$ as the $0^{th}$ detection result). During testing, the outputs from all refinement branches are averaged for the final detection result $\mathbf{D}=\mathbf{D}^r=\frac{1}{3}\sum_{i=1}^3 D^{ri}$.

The segmentation branch $f^S$ is shown in Fig.~\ref{fig:seg} and it is similar to the RefineNet in CPN~\cite{Chen2018CPN}, which is effective in integrating the multi-scale information for the accurate localization problem. As it is illustrated in \cite{Chen2018CPN}, the architecture, mainly consisting of several stacked bottleneck blocks, can transmit information across different scales and integrate all of them. The normalization layers in the bottleneck blocks are changed from the batch normalization to group normalization~\cite{wu2018group} in our experiments, given that the batch size is too small to train a good batch normalization layer.


\section{Appendix: Further Ablation Study}
\label{sec:abl}

\subsection{Investigation of Hyper-parameters}
\begin{figure}[h!]
\begin{center}
\includegraphics[width=0.85\linewidth]{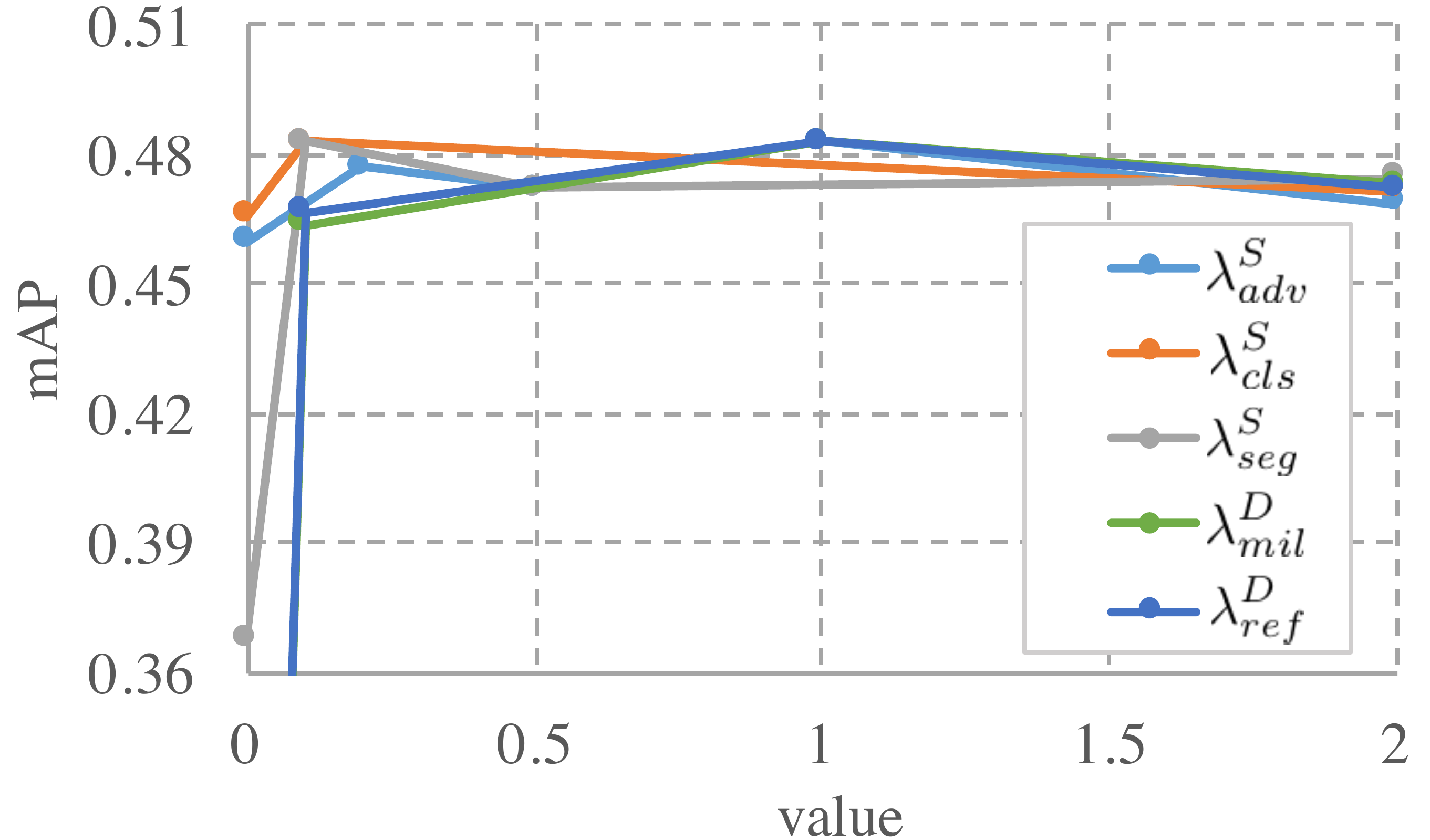}
\end{center}
   \caption{The curves of the mAP varying with the balance weights for each loss on the PASCAL VOC \emph{test} set.}
\label{fig:weights}
\end{figure}

The influences of balance weights $\lambda^S_{adv}$ , $\lambda^S_{cls}$ , $\lambda^S_{seg}$, $\lambda^D_{mil}$ and $\lambda^D_{ref}$ are shown in Fig.~\ref{fig:weights}. As can be seen, the detection performance is not sensitive to these parameters when they are larger than 0.1, demonstrating the robustness of the proposed method.

\begin{figure*}[hb]
\begin{center}
\hspace{0.5cm}\includegraphics[width=0.9\linewidth]{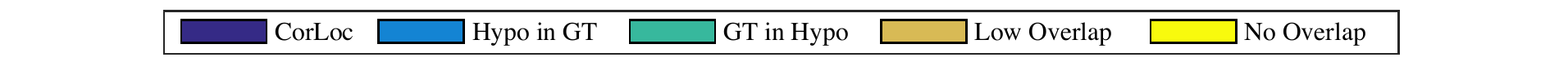}
\subfloat[][OICR]{\includegraphics[width=0.45\linewidth]{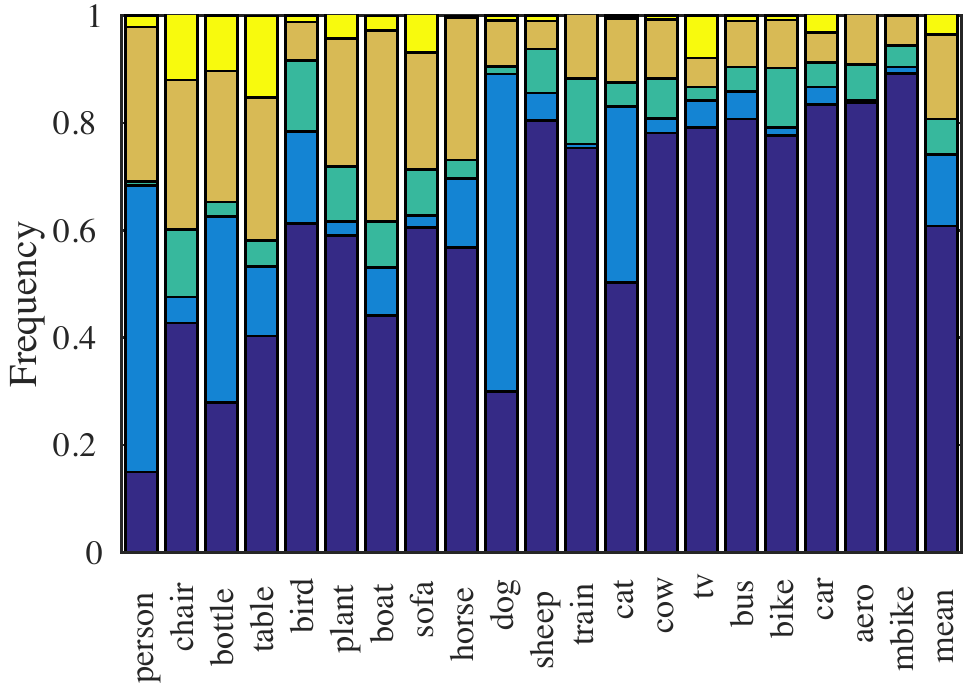}
			   \label{fig:error_oicr}}
\subfloat[][Ours]{\includegraphics[width=0.48\linewidth]{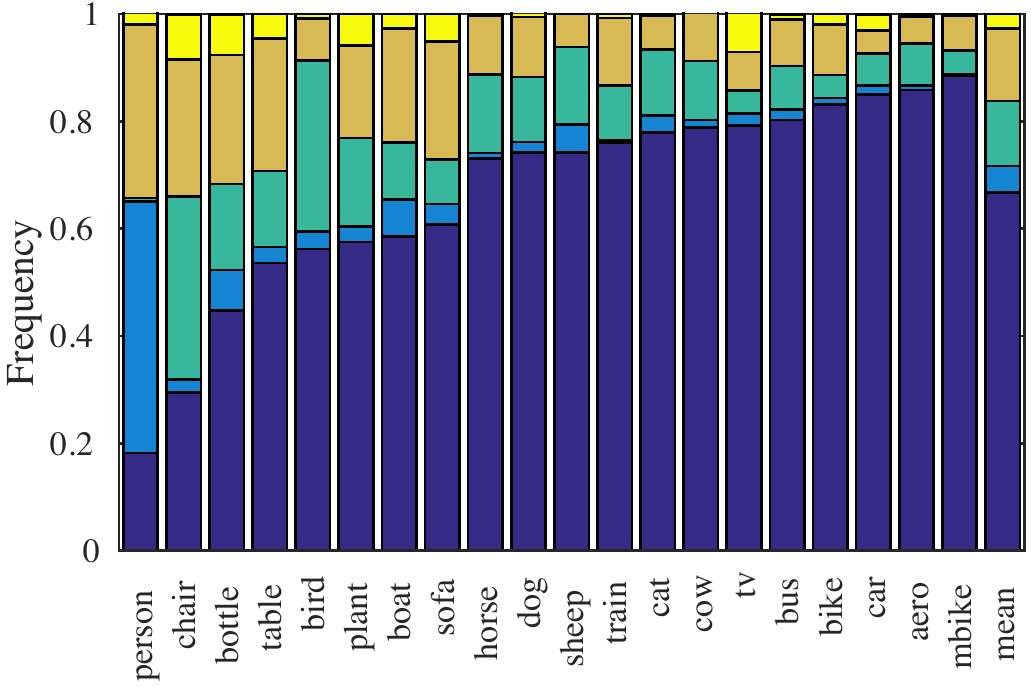}
			   \label{fig:error_sdcn}}
\end{center}
   \caption{Per-class frequencies of error modes, and averaged across all classes for the baseline OICR~\cite{tang2017multiple} and our proposed method on the PASCAL VOC 2007 \emph{trainval} set.}
\label{fig:error}
\end{figure*}




\subsection{Error Analysis}

We investigate the detailed sources of errors following~\cite{cinbis2017weakly}, where detected boxes are categorized into five cases: 1) correct localization (overlap with the ground-truth $\geqslant$ 50\%), 2) the hypothesis completely inside the ground-truth, 3) the ground-truth completely inside the hypothesis, 4) none of the above, but non-zero overlap, and 5) no overlap.

The frequencies of these five cases are shown in Fig.~\ref{fig:error_oicr} for the baseline OICR. The largest error lies in the low overlap between the hypothesis and the ground-truth, which is inevitable for all existing weakly supervised object detector, resulting from hard cases or self-limitations of the detector. It is noticeable that the hypothesis inside the ground-truth is the second largest error mode, which indicates that the OICR frequently confuses object parts with real objects.

 \begin{figure*}[ht]
 	\begin{center}
 	\includegraphics[width=0.95\linewidth]{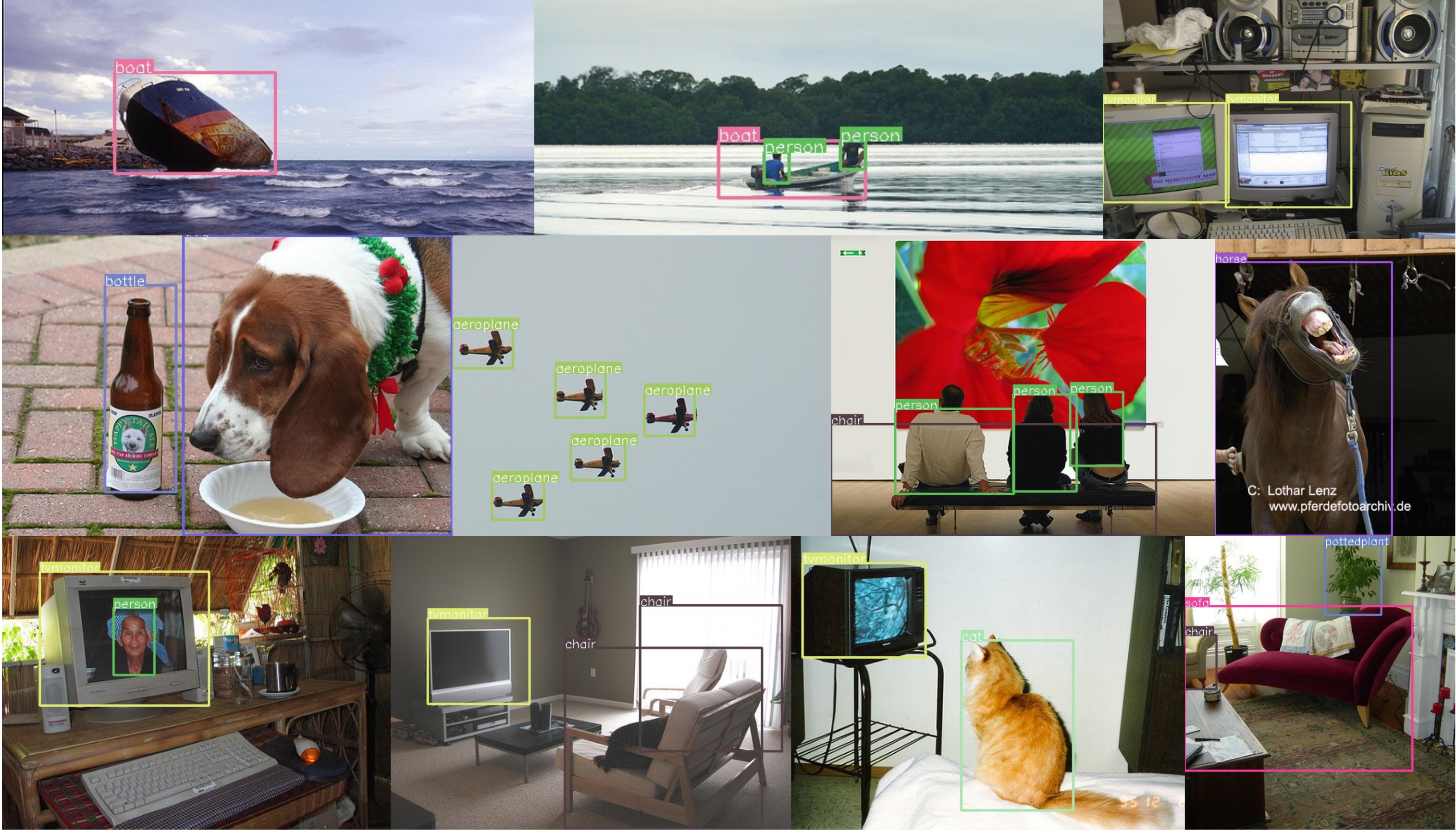}
	\end{center}
	\caption{Visualization of the proposed SDCN on the PASCAL VOC 2007 \emph{test} set.}
	\label{fig:vis}
\end{figure*}

The corresponding result for the proposed SDCN model is shown in Fig.~\ref{fig:error_sdcn}. The area of deep blue bars, representing the ratio of correct localization, increases obviously, and the frequencies for three types of errors decreases, especially for the hypothesis inside the ground-truth. It indicates our method greatly overcomes the mentioned confusion in OICR. However, the cases for ground-truth inside the hypothesis increase inevitably, owing to the utilization of semantic segmentation maps rather than instance segmentation maps, which will be considered in our future work.

Additional visualization of the detection results is shown in Fig.\ref{fig:vis}. As can be seen, although these input images include hard samples, \eg occluded or distorted objects and multiple instances in one image, the proposed method still detect these objects.

{\small
\bibliographystyle{ieee}
\bibliography{egbib}

\begin{thebibliography}{10}\itemsep=-1pt

\bibitem{bilen2014weakly}
H.~Bilen, M.~Pedersoli, and T.~Tuytelaars.
\newblock Weakly supervised object detection with posterior regularization.
\newblock In {\em British Machine Vision Conference (BMVC)}, pages 1--12, 2014.

\bibitem{bilen2016weakly}
H.~Bilen and A.~Vedaldi.
\newblock Weakly supervised deep detection networks.
\newblock In {\em IEEE Conference on Computer Vision and Pattern Recognition
  (CVPR)}, pages 2846--2854, 2016.

\bibitem{blaschko2010simultaneous}
M.~Blaschko, A.~Vedaldi, and A.~Zisserman.
\newblock Simultaneous object detection and ranking with weak supervision.
\newblock In {\em Advances in Neural Information Processing Systems (NIPS)},
  pages 235--243, 2010.

\bibitem{Chen2018CPN}
Y.~Chen, Z.~Wang, Y.~Peng, Z.~Zhang, G.~Yu, and J.~Sun.
\newblock {Cascaded Pyramid Network for Multi-Person Pose Estimation}.
\newblock In {\em IEEE Conference on Computer Vision and Pattern Recognition
  (CVPR)}, 2018.

\bibitem{cinbis2017weakly}
R.~G. Cinbis, J.~Verbeek, and C.~Schmid.
\newblock Weakly supervised object localization with multi-fold multiple
  instance learning.
\newblock {\em IEEE Transactions on Pattern Analysis \& Machine Intelligence
  (TPAMI)}, 39(1):189--203, 2017.

\bibitem{deselaers2010localizing}
T.~Deselaers, B.~Alexe, and V.~Ferrari.
\newblock Localizing objects while learning their appearance.
\newblock In {\em European Conference on Computer Vision (ECCV)}, pages
  452--466, 2010.

\bibitem{deselaers2012weakly}
T.~Deselaers, B.~Alexe, and V.~Ferrari.
\newblock Weakly supervised localization and learning with generic knowledge.
\newblock {\em International Journal of Computer Vision (IJCV)},
  100(3):275--293, 2012.

\bibitem{diba2017weakly}
A.~Diba, V.~Sharma, A.~M. Pazandeh, H.~Pirsiavash, and L.~Van~Gool.
\newblock Weakly supervised cascaded convolutional networks.
\newblock In {\em IEEE Conference on Computer Vision and Pattern Recognition
  (CVPR)}, page~9, 2017.

\bibitem{dietterich1997solving}
T.~G. Dietterich, R.~H. Lathrop, and T.~Lozano-P{\'e}rez.
\newblock Solving the multiple instance problem with axis-parallel rectangles.
\newblock {\em Artificial Intelligence.}, 89(1-2):31--71, 1997.

\bibitem{Durand2017WILDCAT}
T.~Durand, T.~Mordan, N.~Thome, and M.~Cord.
\newblock {WILDCAT: Weakly Supervised Learning of Deep ConvNets for Image
  Classification, Pointwise Localization and Segmentation}.
\newblock In {\em IEEE Conference on Computer Vision and Pattern Recognition
  (CVPR)}, 2017.

\bibitem{everingham2015pascal}
M.~Everingham, S.~A. Eslami, L.~Van~Gool, C.~K. Williams, J.~Winn, and
  A.~Zisserman.
\newblock The pascal visual object classes challenge: A retrospective.
\newblock {\em International Journal of Computer Vision (IJCV)},
  111(1):98--136, 2015.

\bibitem{everingham2010pascal}
M.~Everingham, L.~Van~Gool, C.~K. Williams, J.~Winn, and A.~Zisserman.
\newblock The pascal visual object classes (voc) challenge.
\newblock {\em International Journal of Computer Vision (IJCV)},
  88(2):303--338, 2010.

\bibitem{ge2018multievidence}
W.~Ge, S.~Yang, and Y.~Yu.
\newblock Multi-evidence filtering and fusion for multi-label classification,
  object detection and semantic segmentation based on weakly supervised
  learning.
\newblock In {\em IEEE Conference on Computer Vision and Pattern Recognition
  (CVPR)}, 2018.

\bibitem{girshick2015fast}
R.~Girshick.
\newblock Fast r-cnn.
\newblock In {\em IEEE International Conference on Computer Vision (ICCV)},
  pages 1440--1448, 2015.

\bibitem{gokberk2014multi}
R.~Gokberk~Cinbis, J.~Verbeek, and C.~Schmid.
\newblock Multi-fold mil training for weakly supervised object localization.
\newblock In {\em IEEE Conference on Computer Vision and Pattern Recognition
  (CVPR)}, pages 2409--2416, 2014.

\bibitem{he2015deep}
K.~He, X.~Zhang, S.~Ren, and J.~Sun.
\newblock Deep residual learning for image recognition.
\newblock In {\em IEEE Conference on Computer Vision and Pattern Recognition
  (CVPR)}, 2016.

\bibitem{jie2017deep}
Z.~Jie, Y.~Wei, X.~Jin, J.~Feng, and W.~Liu.
\newblock Deep self-taught learning for weakly supervised object localization.
\newblock In {\em IEEE Conference on Computer Vision and Pattern Recognition
  (CVPR)}, 2017.

\bibitem{kolesnikov2016seed}
A.~Kolesnikov and C.~H. Lampert.
\newblock Seed, expand and constrain: Three principles for weakly-supervised
  image segmentation.
\newblock In {\em European Conference on Computer Vision (ECCV)}, 2016.

\bibitem{russakovsky2012object}
O.~Russakovsky, Y.~Lin, K.~Yu, and L.~Fei-Fei.
\newblock Object-centric spatial pooling for image classification.
\newblock In {\em European Conference on Computer Vision (ECCV)}, pages 1--15,
  2012.

\bibitem{shen2018generative}
Y.~Shen, R.~Ji, S.~Zhang, W.~Zuo, and Y.~Wang.
\newblock Generative adversarial learning towards fast weakly supervised
  detection.
\newblock In {\em IEEE Conference on Computer Vision and Pattern Recognition
  (CVPR)}, 2018.

\bibitem{simonyan14very}
K.~Simonyan and A.~Zisserman.
\newblock Very deep convolutional networks for large-scale image recognition.
\newblock In {\em International Conference on Learning Representations (ICLR)},
  2014.

\bibitem{siva2012defence}
P.~Siva, C.~Russell, and T.~Xiang.
\newblock In defence of negative mining for annotating weakly labelled data.
\newblock In {\em European Conference on Computer Vision (ECCV)}, pages
  594--608, 2012.

\bibitem{siva2011weakly}
P.~Siva and T.~Xiang.
\newblock Weakly supervised object detector learning with model drift
  detection.
\newblock In {\em IEEE International Conference on Computer Vision (ICCV)},
  pages 343--350, 2011.

\bibitem{song2014on}
H.~O. Song, R.~Girshick, S.~Jegelka, J.~Mairal, Z.~Harchaoui, and T.~Darrell.
\newblock On learning to localize objects with minimal supervision.
\newblock In {\em International Conference on Machine Learning (ICML)}, pages
  1611--1619, 2014.

\bibitem{song2014weakly}
H.~O. Song, Y.~J. Lee, S.~Jegelka, and T.~Darrell.
\newblock Weakly-supervised discovery of visual pattern configurations.
\newblock In {\em Advances in Neural Information Processing Systems (NIPS)},
  pages 1637--1645, 2014.

\bibitem{tang2017multiple}
P.~Tang, X.~Wang, X.~Bai, and W.~Liu.
\newblock Multiple instance detection network with online instance classifier
  refinement.
\newblock In {\em IEEE Conference on Computer Vision and Pattern Recognition
  (CVPR)}, 2017.

\bibitem{tang2018weakly}
P.~Tang, X.~Wang, A.~Wang, Y.~Yan, W.~Liu, J.~Huang, and A.~Yuille.
\newblock Weakly supervised region proposal network and object detection.
\newblock In {\em European Conference on Computer Vision (ECCV)}, 2018.

\bibitem{uijlings2013selective}
J.~R.~R. Uijlings, K.~E. A.~V. De~Sande, T.~Gevers, and A.~W.~M. Smeulders.
\newblock Selective search for object recognition.
\newblock {\em International Journal of Computer Vision (IJCV)},
  104(2):154--171, 2013.

\bibitem{wan2018min}
F.~Wan, P.~Wei, J.~Jiao, Z.~Han, and Q.~Ye.
\newblock Min-entropy latent model for weakly supervised object detection.
\newblock In {\em IEEE Conference on Computer Vision and Pattern Recognition
  (CVPR)}, pages 1297--1306, 2018.

\bibitem{wei2017advers}
Y.~Wei, J.~Feng, X.~Liang, M.-M. Cheng, Y.~Zhao, and S.~Yan.
\newblock Object region mining with adversarial erasing: A simple
  classification to semantic segmentation approach.
\newblock In {\em IEEE Conference on Computer Vision and Pattern Recognition
  (CVPR)}, 2017.

\bibitem{wei2018ts2c}
Y.~Wei, Z.~Shen, B.~Cheng, H.~Shi, J.~Xiong, J.~Feng, and T.~Huang.
\newblock Ts2c: tight box mining with surrounding segmentation context for
  weakly supervised object detection.
\newblock In {\em European Conference on Computer Vision (ECCV)}, 2018.

\bibitem{wu2018group}
Y.~Wu and K.~He.
\newblock Group normalization.
\newblock In {\em European Conference on Computer Vision (ECCV)}, 2018.

\bibitem{zhang2018zigzag}
X.~Zhang, J.~Feng, H.~Xiong, and Q.~Tian.
\newblock Zigzag learning for weakly supervised object detection.
\newblock In {\em IEEE Conference on Computer Vision and Pattern Recognition
  (CVPR)}, 2018.

\bibitem{zhou2018a}
Z.-H. Zhou.
\newblock A brief introduction to weakly supervised learning.
\newblock {\em National Science Review}, 5(1):44--53, 2018.

\end{thebibliography}
}

\end{document}